\documentclass[default,iicol]{sn-jnl}

\jyear{2023}%

\theoremstyle{thmstyleone}%
\theoremstyle{thmstyletwo}%

\theoremstyle{thmstylethree}%

\usepackage{subcaption}

\begin{document}

\title[Image Models and Tablets]{CNN-based Image Models Verify a Hypothesis that The Writers of Cuneiform Texts Improved Their Writing Skills When Studying at the Age of Hittite Empire}


\author*[1]{\fnm{Daichi} \sur{Kohmoto}}\email{kohmoto.daichi@me.com}

\author[1]{\fnm{Katsutoshi} \sur{Fukuda}}

\author[2]{\fnm{Daisuke} \sur{Yoshida}}

\author[3]{\fnm{Takafumi} \sur{Matsui}}

\author[2]{\fnm{Sachihiro} \sur{Omura}}

\affil*[1]{\orgdiv{Office of Society-Academia Collaboration for Innovation}, \orgname{Kyoto University}, \orgaddress{\street{Yoshida-honmachi, Sakyo-ku}, \city{Kyoto}, \postcode{6068501}, \state{Kyoto}, \country{JAPAN}}}

\affil[2]{\orgdiv{Japanese Institute of Anatolian Archaeology}, \orgname{The Middle Eastern Culture Center in Japan}, \orgaddress{\street{3-10-31 Osawa}, \city{Mitaka}, \postcode{1810015}, \state{Tokyo}, \country{JAPAN}}}

\affil[3]{\orgname{Chiba Institute of Technology}, \orgaddress{\street{2-17-1 Tsudanuma}, \city{Narashino}, \postcode{2750016}, \state{Chiba}, \country{JAPAN}}}

\abstract{
A cuneiform tablet KBo 23.1 ++/KUB 30.38, which is known to represent a text of Kizzuwatna rituals, was written by two writers with almost identical content in two iterations. Unlike other cuneiform tablets that contained information such as myths, essays, or business records, the reason why ancient people left such tablets for posterity remains unclear. To study this problem, we develop a new methodology by analyzing images of a tablet quantitatively using CNN (Convolutional Neural Network)-based image models, without segmenting cuneiforms one-by-one. Our data-driven methodology implies that the writer writing the first half was a `teacher' and the other writer was a `student' who was training his skills of writing cuneiforms. This result has not been reached by classical linguistics. We also discuss related conclusions and possible further directions for applying our method and its generalizations.
}


\maketitle

\section{Preface}\label{Preface}

Since ancient times, letters have been crucial for supporting civilizations and keeping track of various records and ideas.
Apparently, schools for teaching the writing skills existed in those periods.
For example, in a circular tablet excavated in the Mesopotamian region \cite{another_tablet_NY}, with an estimated date approximately from 2000 BC to 1600 BC, students practiced in writing the name of the God Urash in Sumerian and Akkadian in cuneiform six times.
Common theories and hypotheses assume that reading and writing were popular among officials and military personnels in this era, and that young potential candidates practised their reading and writing skills at schools, called Edubba in ancient Mesopotamia. 
A potential Edubba was found in the royal palace of Zimri-Lim in the Mali ruins of Sumerian civilization (e.g., \cite{edubba1}, \cite{edubba2}, and \cite{edubba3}). 
The first key to discovering the existence of such schools would be finding clay tablets showing the evidence of training the skills of writing cuneiforms. 

A unique clay tablet KBo 23.1 ++/KUB 30.38 \cite{CTH} was found in the Hittite capital city, Hattusa (see Fig. \ref{fig:rawimages}) and belongs to the period of the Hittite Empire (1400 BC to 1300 BC). 
It is known as a document about the rituals of Kizzuwatna, a region in South-East Turkey under the Hittite control from the fifteenth century BC during the reign of the King Tuthaliya I/II, \cite{Rita} and \cite{review_CTH472}. 
From linguistic viewpoint, it is known that two authors wrote almost identical sentences in this tablet, and a former work suggests this tablet might be evidence of `a pedagogic exercise' concerning one of those two authors \cite{Yoshida}. 
These motivate us to study this tablet as potential evidence of practising skills of writing cuneiforms, via handwriting analysis. 

This study incorporates machine learning, which has been rapidly developing in recent years, into the conventional linguistics methodologies. Machine learning makes precise predictions based on large-scale data with fairness, reproductivity, and explainability. 
In this study, our data analysis is performed using three 2D image data of the clay tablet KBo 23.1 ++/KUB 30.38 in Hethitologie-Portals Mainz \cite{CTH}. 

Many previous studies focused on identifying hundreds order of authors by analyzing handwritten characters using machine learning for modern languages such as English, Spanish, French, and Arabic.
These research projects are collectively called the {\it writer-identification problem}.
A comprehensive review \cite{review_wiriteridentification} describes the research on writer identification using machine learning. 
As the wedge-shaped characters written in a tablet are handwritten characters, the evidence of improvements in training skills of writing cuneiforms might be obtained quantitatively via these technologies in addition to identifying the authors. 
This study has two parts: handwriting analysis and writer identification analysis. 


\section{Methodology and Experiments}\label{Methodology}
We start from extracting rectangular image pieces of two-line sentences from three images of our target tablet KBo 23.1 ++/KUB 30.38 as seen in Fig. \ref{fig:rawimages}. Next, we define $n$ classes for these pieces assuming $n$ candidate writers, and construct the main datasets enabling $5$-fold validation via cropping squares and data augmentation techniques. 
Almost as in some cases of writer-identification problems, we model our problem as multi-class classification with multiple CNN-based image models VGG19 \cite{vgg19}, ResNet50 \cite{resnet50}, and InceptionV3 \cite{inceptionv3}, trained on ImageNet \cite{imagenet}, which is a well-known large-scale dataset of images. 
Differences between writer-identification problems and our approach are (1) the way of constructing the datasets for building our models, because the number of images of our target tablet is limited (e.g., three images), and writers are not true writers but `candidate' writers in the first place,  (2) that we focus not only on achieving state-of-the-art precision when we fine-tune image models, but also on non-matching parts between true labels and predictions of its models for our analysis, which we will define as {\it similarity}, and (3) that we use multiple CNN-based image models and derive conclusions by {\it majority vote} from the results of (1) and (2). 
As we can discuss similarity between $n$ candidate writers from (2) and (3), leads to new insights related to this tablet. Similarity and majority vote, explained in subsections \ref{subsection:howweargue} and \ref{subsection:howweargue_2}, are two main building blocks of our methodology in this study. 

To our knowledge, there are few studies on developing methodologies focusing on specific problems of humanities with only a few images using machine learning. As an aspect of handwriting analysis, there exist a deductive approach \cite{10.1371/journal.pone.0243039} developing methodologies to identify each cuneiform character one-by-one using image datasets of tablets and their transliterations. Since their purposes do not include extracting features of cuneiforms on tablets, one cannot directly apply their methodologies to our problems. Previous research on writer-identification problems \cite{10.1007/s11042-018-6577-1} is focusing on identifying authors in the order of hundreds via developed and optimized one model, so that approaches of our study are quite different from them.

This study uses pretrained PyTorch Image Models {\tt timm} \cite{rw2019timm} as CNN-based image models. 
We fine-tune these pretrained models over augmented main datasets enabling $5$-fold validations, and check (i) the results of confusion matrices that we obtain by applying fine-tuned models to the test parts of the main datasets, (ii) the results of external validations by applying fine-tuned models to other datasets built from two additional parts of the target tablet to be identified, and (iii) the results of class activation mapping to understand what features of image samples in the main datasets these models encode for identifying the class of the image sample.

\subsection{Building Main Datasets}\label{augmentation}
\begin{figure}[htbp]
\centering
  \begin{minipage}[b]{0.3\linewidth}
    \centering
    \includegraphics[height=3.1cm]{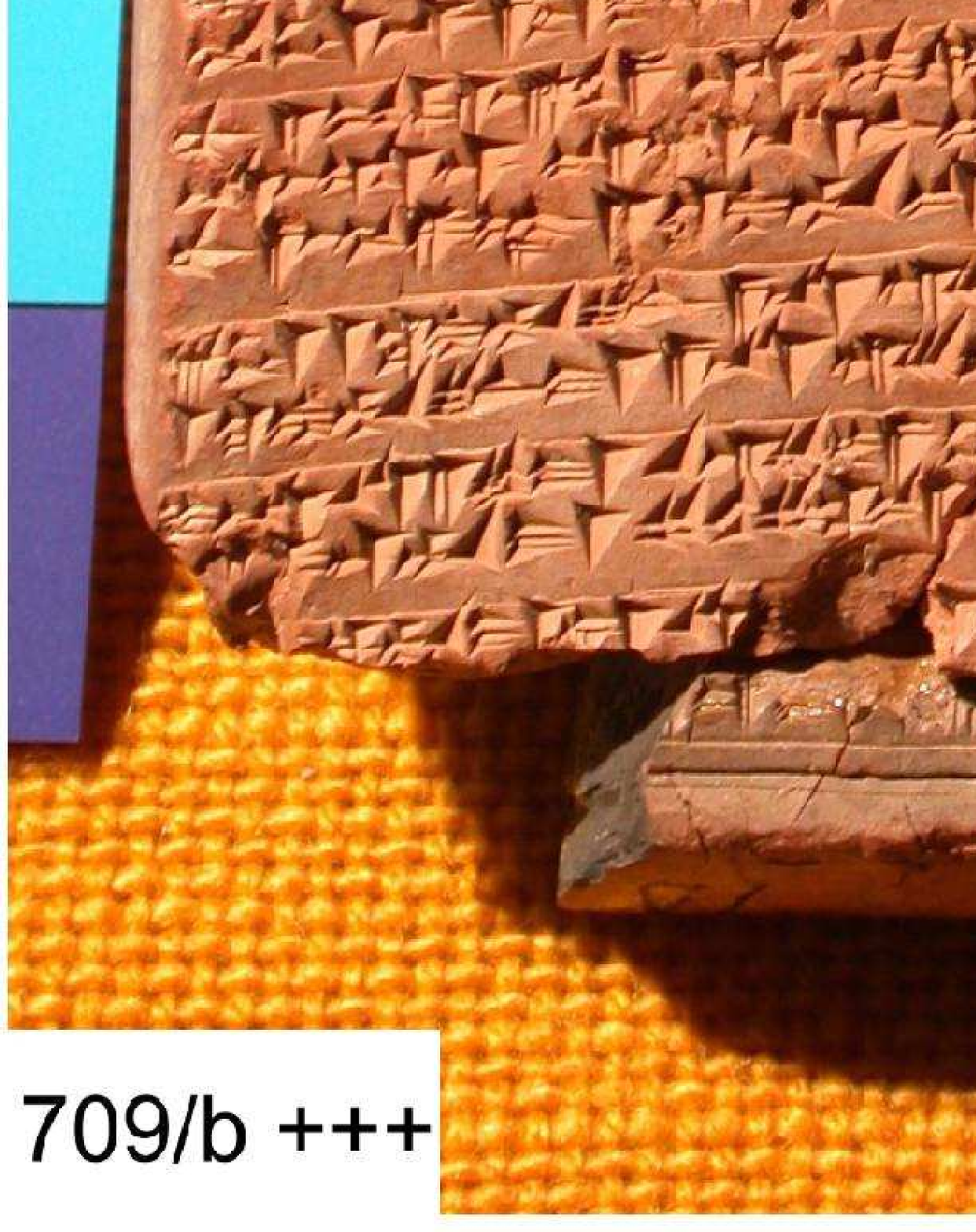}
    \label{fig:frontside1}
  \end{minipage}
  \begin{minipage}[b]{0.3\linewidth}
    \centering
    \includegraphics[height=3.1cm]{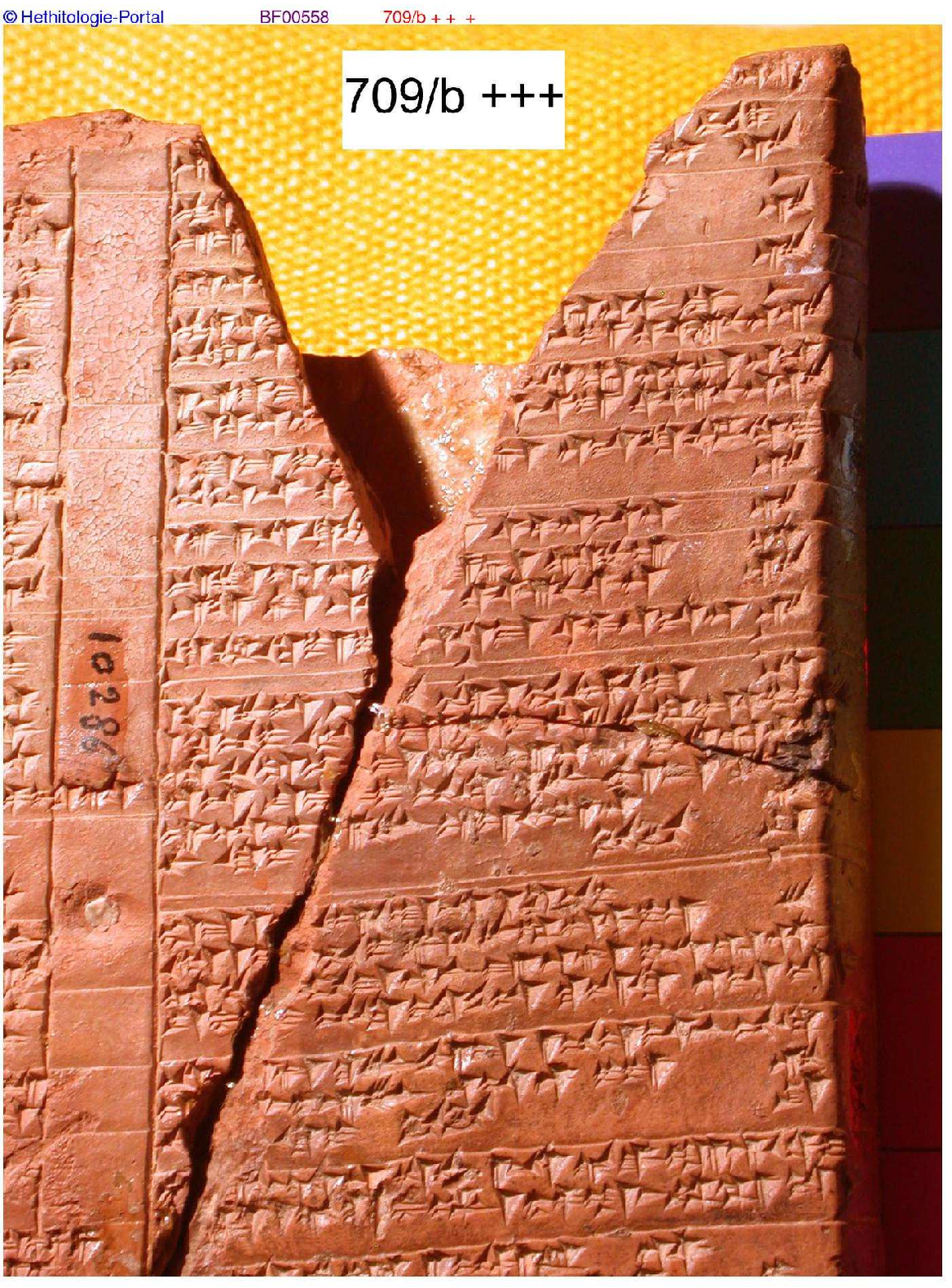}
    \label{fig:frontside2}
  \end{minipage}
  \begin{minipage}[b]{0.3\linewidth}
    \centering
    \includegraphics[height=3.1cm]{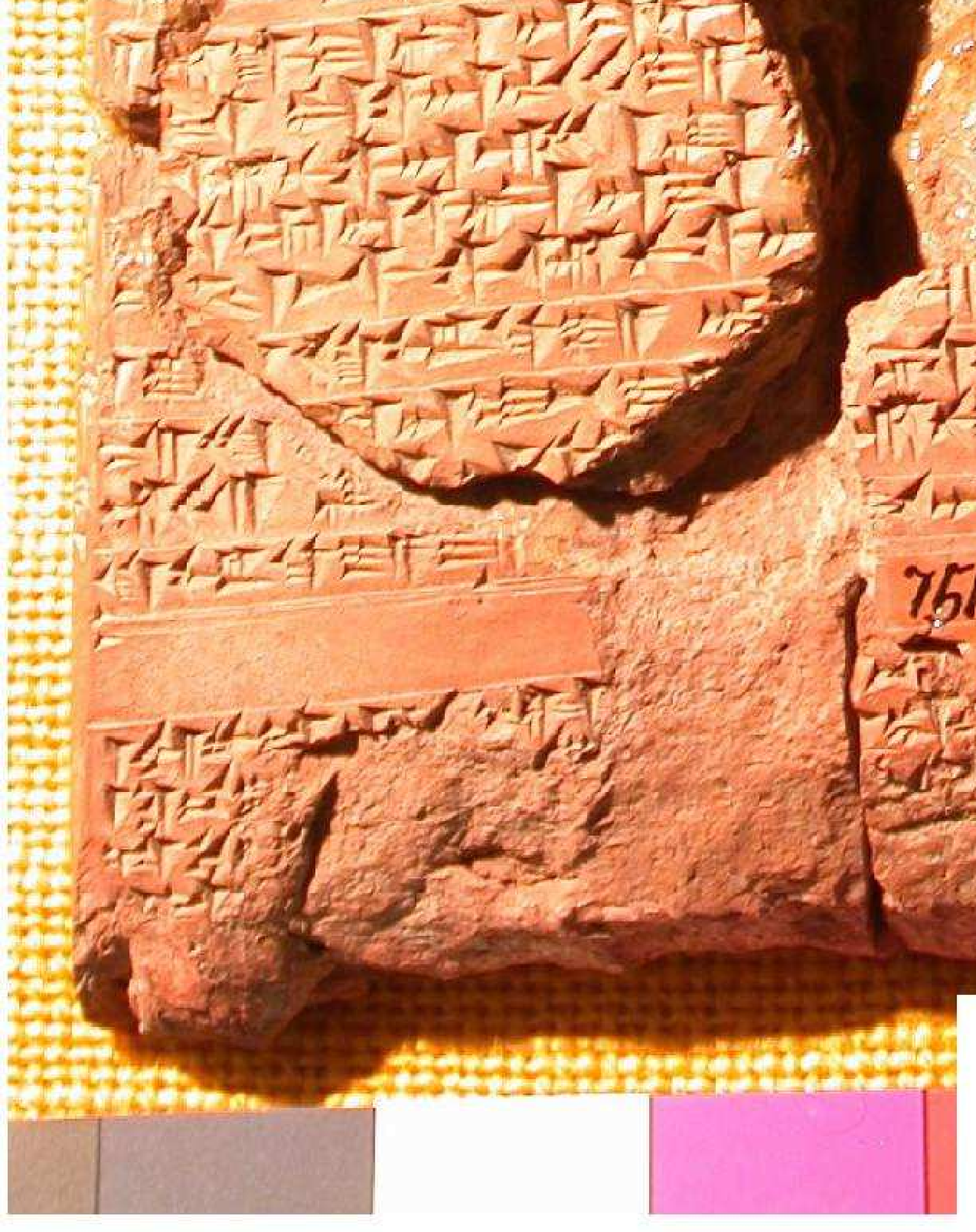}
    \label{fig:backside}
  \end{minipage}
  \caption{Raw Images of Target Tablet KBo 23.1 ++/KUB 30.38: (a) Left-Bottom Part of Front Side 1, (b) Right-Top Part in Front Side 2, and (c) Back Side}
  \label{fig:rawimages}
\end{figure}

Our target tablet KBo 23.1 ++/KUB 30.38 have two-sided columns of sentences, as shown in Fig. \ref{fig:rawimages}: this study borrows and uses these three original pictures from \cite{CTH}. 
According to the current understanding of linguistics, two authors wrote the same sentences in the tablet:  
due to the existence of certain signs at the end of the tablet, the first column and almost half of the second column (right half of the front side) up to the 22nd line of the front side were written by one author, and the remaining parts of the tablet, including the back side, were written by the other author.

\begin{figure}[htbp]
	\centering
	\begin{minipage}[b]{0.3\linewidth}
		\centering
		\includegraphics[height=3.1cm]{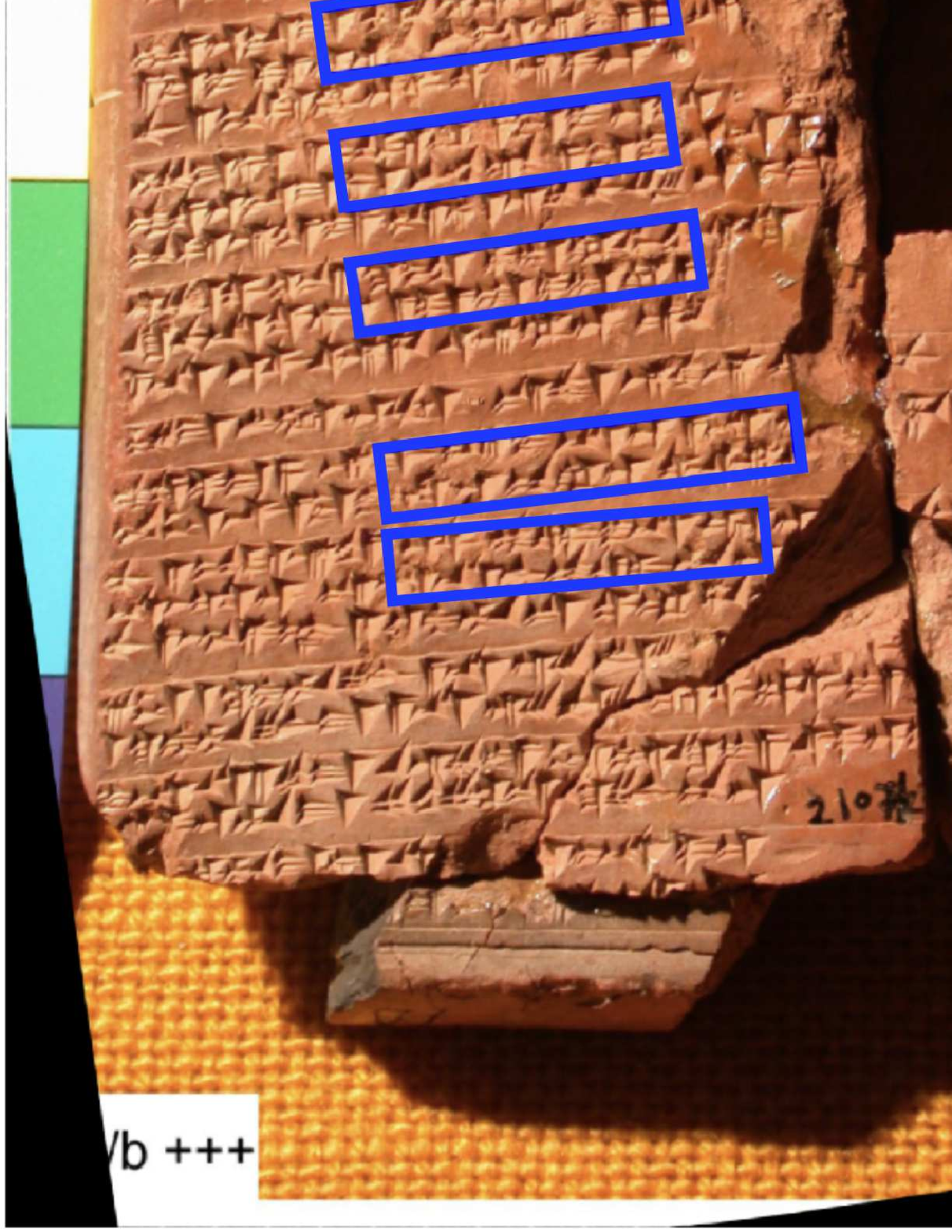}
		\label{fig:class1of4}
	\end{minipage}
	\begin{minipage}[b]{0.3\linewidth}
		\centering
		\includegraphics[height=3.1cm]{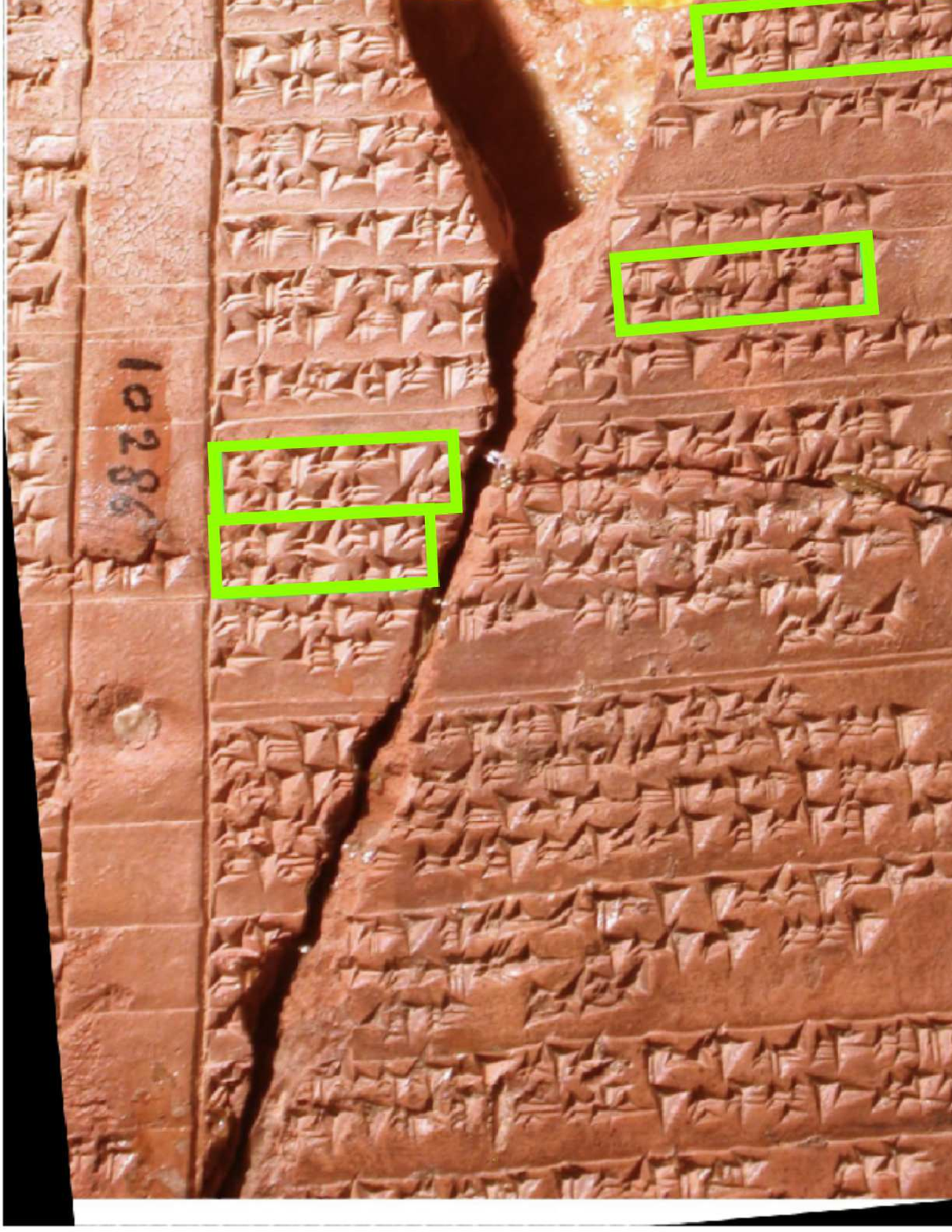}
		\label{fig:class2of4}
	\end{minipage}
	\begin{minipage}[b]{0.3\linewidth}
		\centering
		\includegraphics[height=3.1cm]{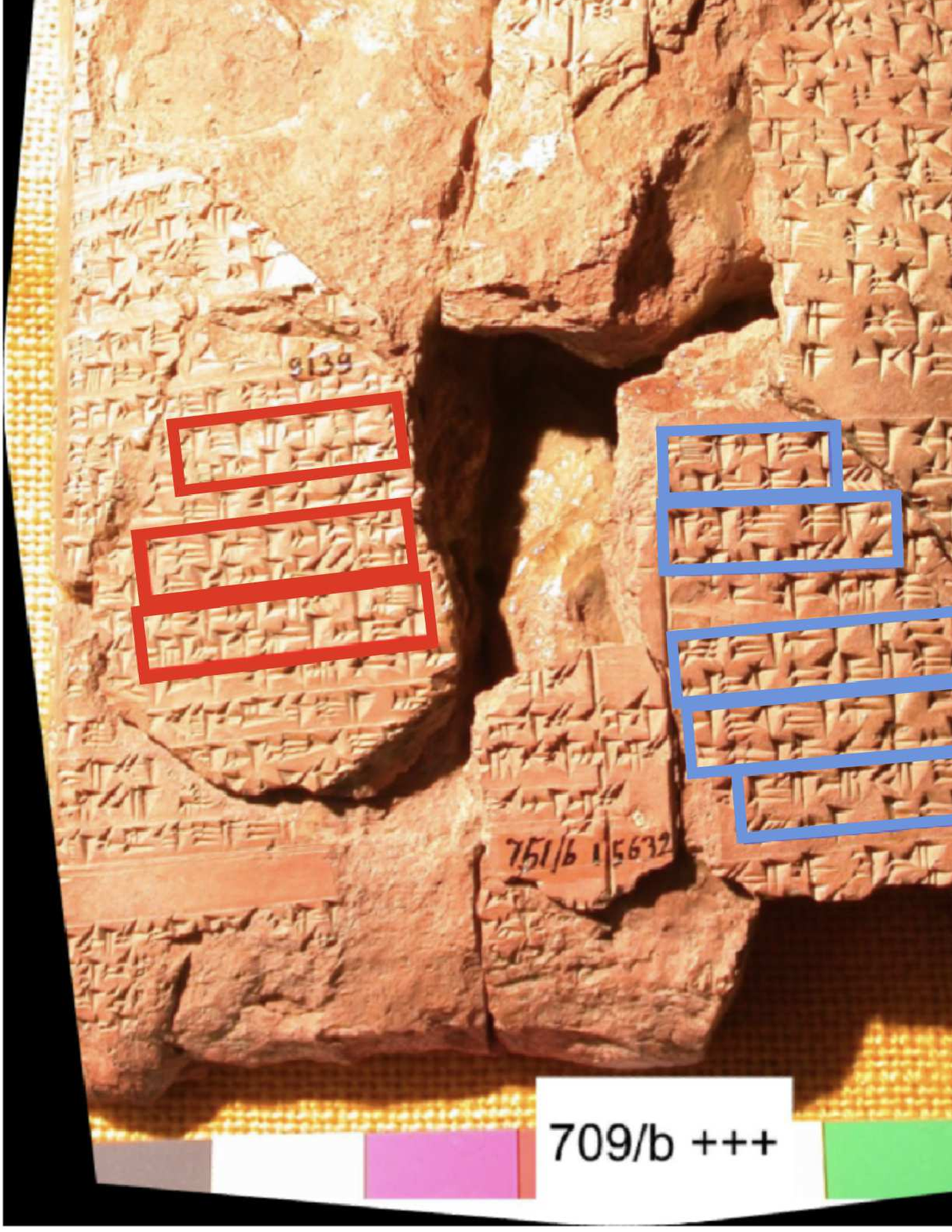}
		\label{fig:class34of4}
	\end{minipage}\\

	\centering
	\begin{minipage}[b]{0.3\linewidth}
		\centering
		\includegraphics[height=3.1cm]{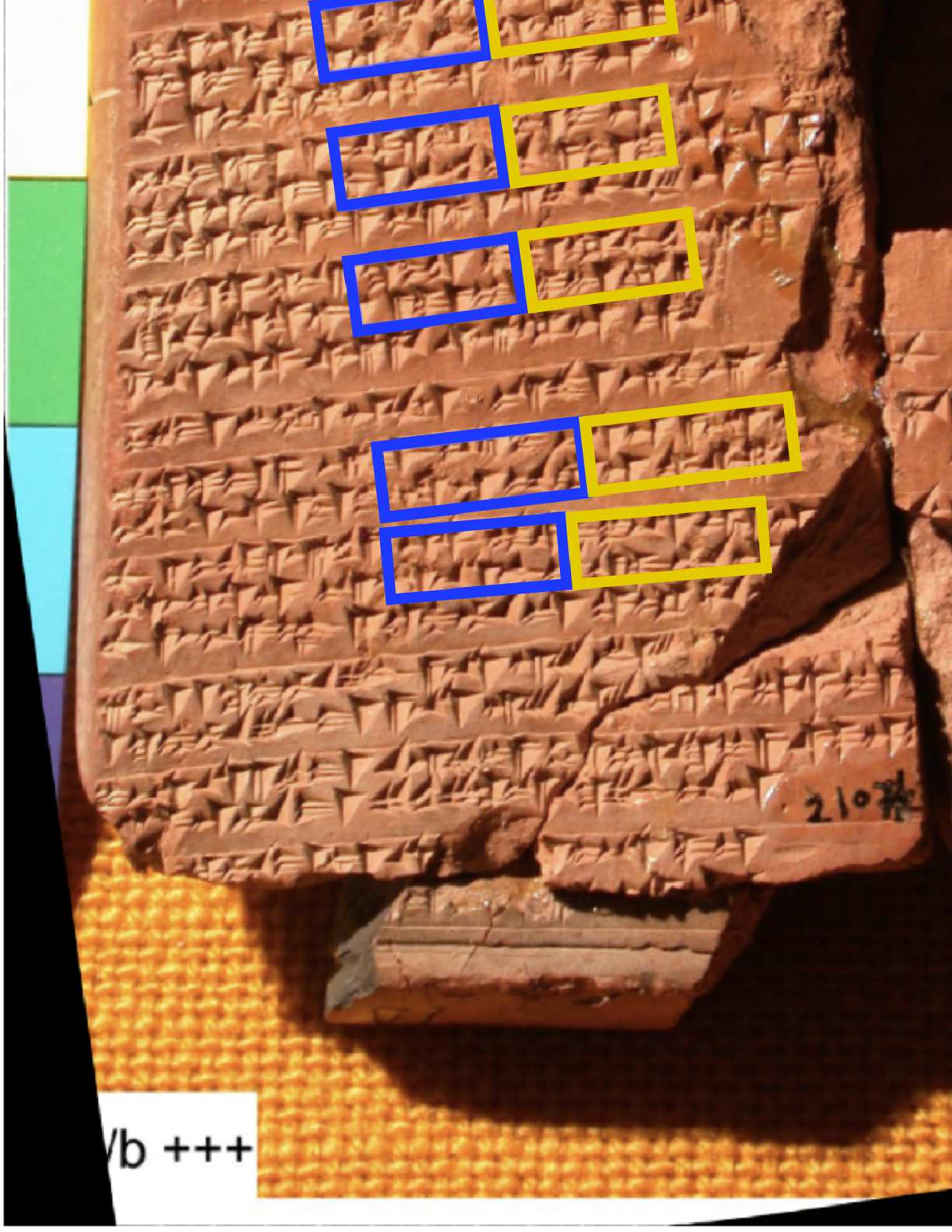}
		\label{fig:class12of8}
	\end{minipage}
	\begin{minipage}[b]{0.3\linewidth}
		\centering
		\includegraphics[height=3.1cm]{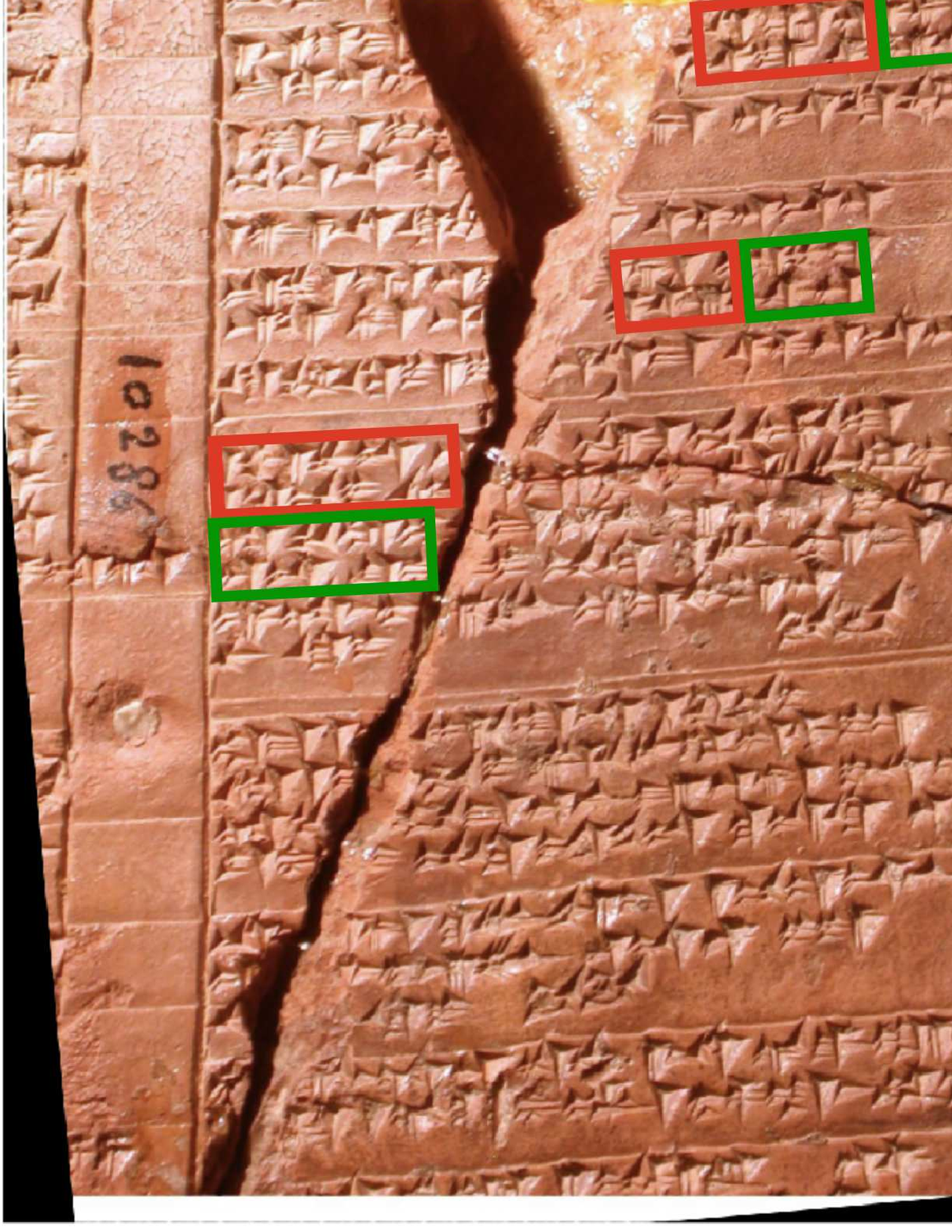}
		\label{fig:class34of8}
	\end{minipage}
	\begin{minipage}[b]{0.3\linewidth}
		\centering
		\includegraphics[height=3.1cm]{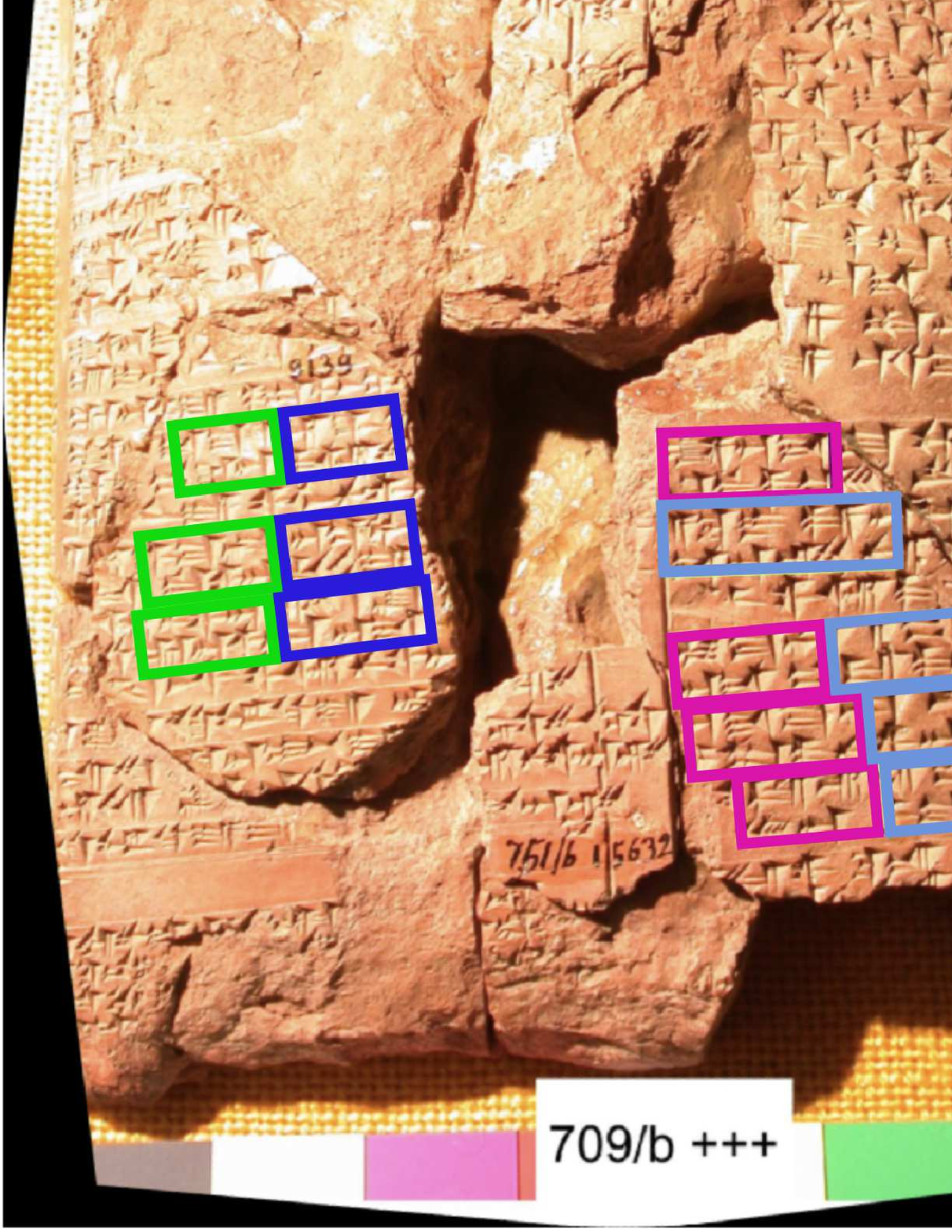}
		\label{fig:class5678of8}
	\end{minipage}
\caption{Definition of Classes: (a) Class $1$ in $4$ classes, (b) Class $2$ in $4$ classes, (c) Class $3$ \& $4$ in $4$ classes, (d) Class $1,2$ in $8$ classes, (e) Class $3,4$ in $8$ classes, and (f) Class $5$-$8$ in $8$ classes}
\label{fig:defclasses}
\end{figure}

We crop rectangular image pieces of two-line cuneiform sentences as a first step. 
The original images used in this study contain various biasses, including uneven sheen and non-flatness of the tablet. 
Therefore, we chose appropriate parts of cuneiform sentences to crop for our study, shown in Fig. \ref{fig:defclasses}.
Note that (1) the contrast between images is not adjusted because there exist an operation ``shining'' in the procedure of data augmentation corresponding the alignment in this study, and (2) we sometimes crop image pieces at a slight angle along their line so that writing direction becomes horizontal. 

Two sets of classes are defined for rectangular image pieces for our analysis: $n=4$ classes and $n=8$ classes. 
Colored boxes corresponding to $4$ classes are displayed in Fig.s \ref{fig:defclasses} a-c: 
blue, yellow-green, red, and light blue boxes represent Class $1,2,3,$ and $4$, respectively. 
The case of $8$ classes is shown in Fig.s \ref{fig:defclasses} d-f: 
blue, yellow, red, green, light green, dark blue, pink, and gray-blue boxes represent 
Class $1,2,3,4,5,6,7,$ and $8$, respectively. 

Roughly speaking, the usage of $4$ classes corresponds to a {\it macroscopic} analysis of images, whereas the usage of $8$ classes corresponds to a {\it microscopic} analysis of images. 

To build our training and testing datasets, we split all rectangular image pieces into training and test parts at first, and finally we augment and crop them into square images. 
Scikit-learn {\tt train\_test\_split} is used for splitting rectangular image pieces five times with split ratio $0.8$ via five prescribed random seeds for validation: $1033$, $1931$, $2201$, $4179$, and $9325$. 
We adopt the following $5$ or $6$ operations for augmentation: ``shining,'' ``horizontal shift,'' ``vertical shift,'' ``horizontal flip,'' ``vertical flip,'' and optionally ``zoom.'' 
The ``zoom'' operation is used to eliminate characteristics caused by the size of the letters written on the tablet from our analysis. 
We set the size of square images for cropping as minimum of all augmented rectangular images, and we crop square images as much as possible from augmented rectangular image pieces by shifting 10 or 20 pixels horizontally and vertically. 
A dataset type is defined for a triple of a zoom option, a shifting value, and a case of the number of classes as v01-v004. 
Each dataset can be characterized by a dataset type and a seed. 

As a result, $40$ ``Main Datasets'' are obtained, as listed in Table \ref{table:datasetstats}, over two kinds of classes ($4$ or $8$), five seeds ($1033$, $1931$, $2201$, $4179$, and $9325$), two augmentation options (using ``zoom'' or not), and two shifting-pixel values ($10$ or $20$). 
Numerical results obtained in the analysis of these $40$ datasets are used to derive a final result by {\it majority vote}.

\begin{table*}[h]
\begin{center}
\begin{minipage}{\textwidth}
\caption{Main Datasets}\label{table:datasetstats}
\begin{tabular*}{\textwidth}{@{\extracolsep{\fill}}ccrrrr@{\extracolsep{\fill}}}
\toprule%
Dataset Type & Zoom Option & Shifting Value of Pixels & \#Samples & \#Classes & Seed \\
\midrule
    v01 & off & $20$ & $508$ & $4$ & $1033$ \\
    v01 & off & $20$ & $508$ & $4$ & $1931$ \\
    v01 & off & $20$ & $508$ & $4$ & $2201$ \\
    v01 & off & $20$ & $508$ & $4$ & $4179$ \\
    v01 & off & $20$ & $508$ & $4$ & $9325$ \\
    v02 & on & $20$ & $8074$ & $4$ & $1033$ \\
    v02 & on & $20$ & $8031$ & $4$ & $1931$ \\
    v02 & on & $20$ & $8078$ & $4$ & $2201$ \\
    v02 & on & $20$ & $8059$ & $4$ & $4179$ \\
    v02 & on & $20$ & $8060$ & $4$ & $9325$ \\
    v03 & off & $10$ & $1546$ & $4$ & $1033$ \\
    v03 & off & $10$ & $1546$ & $4$ & $1931$ \\
    v03 & off & $10$ & $1546$ & $4$ & $2201$ \\
    v03 & off & $10$ & $1546$ & $4$ & $4179$ \\
    v03 & off & $10$ & $1546$ & $4$ & $9325$ \\
    v04 & on & $10$ & $24587$ & $4$ & $1033$ \\
    v04 & on & $10$ & $24433$ & $4$ & $1931$ \\
    v04 & on & $10$ & $24579$ & $4$ & $2201$ \\
    v04 & on & $10$ & $24547$ & $4$ & $4179$ \\
    v04 & on & $10$ & $24555$ & $4$ & $9325$ \\
    v001 & off & $20$ & $428$ & $8$ & $1033$ \\
    v001 & off & $20$ & $428$ & $8$ & $1931$ \\
    v001 & off & $20$ & $428$ & $8$ & $2201$ \\
    v001 & off & $20$ & $428$ & $8$ & $4179$ \\
    v001 & off & $20$ & $428$ & $8$ & $9325$ \\
    v002 & on & $20$ & $6783$ & $8$ & $1033$ \\
    v002 & on & $20$ & $6785$ & $8$ & $1931$ \\
    v002 & on & $20$ & $6785$ & $8$ & $2201$ \\
    v002 & on & $20$ & $6809$ & $8$ & $4179$ \\
    v002 & on & $20$ & $6813$ & $8$ & $9325$ \\
    v003 & off & $10$ & $1294$ & $8$ & $1033$ \\
    v003 & off & $10$ & $1294$ & $8$ & $1931$ \\
    v003 & off & $10$ & $1294$ & $8$ & $2201$ \\
    v003 & off & $10$ & $1294$ & $8$ & $4179$ \\
    v003 & off & $10$ & $1294$ & $8$ & $9325$ \\
    v004 & on & $10$ & $20531$ & $8$ & $1033$ \\
    v004 & on & $10$ & $20517$ & $8$ & $1931$ \\
    v004 & on & $10$ & $20552$ & $8$ & $2201$ \\
    v004 & on & $10$ & $20595$ & $8$ & $4179$ \\
    v004 & on & $10$ & $20584$ & $8$ & $9325$ \\
\bottomrule
\end{tabular*}
\end{minipage}
\end{center}
\end{table*}

\subsection{Study of Similarity between Classes}\label{subsection:howweargue}
After fine-tuning our image models as $n$-class classification tasks, we check not only precisely predicted cases but also non-matching cases between the true labels and predictions on the test datasets using confusion matrices. 
These matrices use $i$-th column and $i$-th row for representing Class $i$. 
The number of non-matching cases between classes can be regarded as an indicator of similarity between classes in this study. 

If we obtain a dummy confusion matrix $A=(a_{i,j})_{0\le i,j \le 4}$ in the case of $4$ classes shown in Table \ref{table:exampleconfusionmatrix} after applying the image model, the matrix has two non-zero non-matching cells: 
$a_{1,2} = 11$ between Class $1$ and Class $2$ and $a_{3,4} = 24$ between Class $3$ and Class $4$. 
The {\it similarity between Class $i$ and $j$} is defined as a summand of $a_{i,j}$ and $a_{j,i}$.  
Accordingly, the similarity $a_{3,4} + a_{4,3} = 24$ between Class $3$ and $4$ is stronger than that $a_{1,2} + a_{2,1} = 11$ between Class $1$ and $2$. 
This viewpoint intends to be used in handwriting analysis, e.g., in subsection \ref{CMresult}, and plays important role in our study.

\begin{table*}[h]
\begin{center}
\begin{minipage}{\textwidth}
\caption{Dummy Example of Confusion Matrix $A$: the Case of $4$ Classes}
\label{table:exampleconfusionmatrix}
\begin{tabular*}{\textwidth}{@{\extracolsep{\fill}}lcccc@{\extracolsep{\fill}}}
\toprule%
& \multicolumn{4}{@{}c@{}}{Predicted Class}  \\\cmidrule{2-5}%
True Class & Class 1 & Class 2 & Class 3 & Class 4 \\
\midrule
Class 1  &  45 & 11 & 0 & 0 \\
Class 2  &  0 & 73 & 0 & 0 \\
Class 3  &  0 & 0 & 31 & 24 \\
Class 4  &  0 & 0 & 0 & 67 \\
\bottomrule
\end{tabular*}
\end{minipage}
\end{center}
\end{table*}

In the case of $8$ classes, similarity is calculated by analogy to the case of $4$ classes, but the focus area and units in a confusion matrix are slightly different. 
Assume we obtain a confusion matrix $B=(b_{i,j})_{0\le i,j \le 8}$ in the case of $8$ classes as in Table \ref{table:exampleconfusionmatrix_2}. 
Class $i$ and Class $i+1$ can be assigned to the same class for an odd integer $0 \leq i \leq 8$ according to the way our main datasets are created. 
We define a ``matching case'' as a square sub-matrix of $B$ consisting of four elements $b_{i,i},b_{i,i+1},b_{i+1,i},$ and $b_{i+1,i+1}$ for an odd integer $0 \leq i \leq 8$, and a ``non-matching case'' can be defined as outside parts of the disjoint unions of matching cases for the aforementioned $i$. 
In this case, the {\it similarity can be defined for a neighboring pair of classes} by analogy to the case of $4$ classes, and in addition, it can be defined as 
\begin{eqnarray*}
(b_{i,j}+b_{i,j+1}+b_{i+1,j}+b_{i+1,j+1})\ \ \ \ \ \ \ \ \\ 
\ \ \ \ \ \ \ \ +(b_{j,i}+b_{j+1,i}+b_{j,i+1}+b_{j+1,i+1})
\end{eqnarray*}
for a pair of neighboring two classes, Class $i$ \& $i+1$ and Class $j$ \& $j+1$, where $i,j$ are odd integers from the non-matching parts of $B$. 
For instance, non-zero non-matching parts consist of $2$ cells in this matrix: $11$ between Class $3$ and Class $6$, and $7$ between Class $8$ and Class $2$. 
In this case, one can say the similarity 
$$(b_{3,5} + b_{3,6} + b_{4,5} + b_{4,6}) + (b_{5,3} + b_{6,3} + b_{5,4} + b_{6,4}) = 11$$
between Class $3$ \& $4$ are  Class $5$ \& $6$ is stronger than that 
$$(b_{1,7} + b_{1,8} + b_{2,7} + b_{2,8}) + (b_{7,1} + b_{8,1} + b_{7,2} + b_{8,2})= 7$$
between Class $1$ \& $2$ and Class $7$ \& $8$.

\begin{table*}[h]
\begin{center}
\begin{minipage}{\textwidth}
\caption{Dummy Example of Confusion Matrix $B$: the Case of $8$ Classes}
\label{table:exampleconfusionmatrix_2}
\begin{tabular*}{\textwidth}{@{\extracolsep{\fill}}lcccccccc@{\extracolsep{\fill}}}
\toprule%
& \multicolumn{8}{@{}c@{}}{Predicted Class}  \\\cmidrule{2-9}%
True Class & Class 1 & Class 2 & Class 3 & Class 4 & Class 5 & Class 6 & Class 7 & Class 8 \\
\midrule
Class 1  & 45 & 11 &  0 & 0 & 0 & 0 & 0 & 0 \\
Class 2  &  6 &  73 &  0 & 0 & 0 & 0 & 0 & 7 \\
Class 3  &  0 &   0 & 31 & 24 & 0 & 0 & 0 & 0 \\
Class 4  &  0 &   0 &  0 & 67 & 0 & 0 & 0 & 0 \\
Class 5  &  0 &   0 &  0 & 0 & 5 & 1 & 0 & 0 \\
Class 6  &  0 &   0 & 11 & 0 & 25 & 11 & 0 & 0 \\
Class 7  &  0 &   0 &  0 & 0 & 0 & 0 & 2 & 0 \\
Class 8  &  0 &   0 &  0 & 0 & 0 & 0 & 5 & 14 \\
\bottomrule
\end{tabular*}
\end{minipage}
\end{center}
\end{table*}

\subsection{Study With a Majority Vote}\label{subsection:howweargue_2}

As one cannot compare results for each type of CNN-based image model (e.g., VGG19, ResNet50, or InceptionV3), 
the majority vote to derive final results in this study is determined as the result of three votes corresponding to VGG19, ResNet50, and InceptionV3. 
In the individual cases per a type of CNN-based image model, the majority vote is conducted over the results in response to a triple of a fine-tuned image model, a dataset type, and a seed. 
Therefore, our majority vote has two steps:  (1) the first vote will be done for a type of CNN-based image model, and (2) the second vote is based on these algorithm's opinions. 
Such ``2-step majority vote'' is used in subsections \ref{CMresult} and \ref{ETresult} for handwriting analysis and writer identification analysis, respectively. 

\subsection{Fine-Tuning CNN-based Image Models}\label{ft}
We fine-tune three commonly used CNN-based image models for our analysis: VGG19, ResNet50, and InceptionV3. 
After downloading pretrained models of {\tt timm} related to VGG19, ResNet50, and InceptionV3, we replace {\tt num\_features} as the number of classes $n$ for each dataset, resize image samples to $224 \times 224$ (for VGG19 and ResNet50) and $299 \times 299$ (for InceptionV3), and then we start training these pretrained models for $50$ epochs with optimizer {\tt Adam}, learning rate $0.0001$, and batch size $16$ with a fixed random seed $1$. 
We train these three models for all $40$ datasets within a day with 4 GPUs of NVIDIA A6000, VRAM 48 GB. 
Note that each fine-tuning case can be characterized by a triple: a pretrained CNN-based image model, a dataset type, and a seed. 

\subsection{Class Activation Mapping (CAM): Explainability of Models}\label{cam}
Explainability of fine-tuned models after applying algorithms of class activation mapping such as Grad-CAM \cite{gradcam}, Eigen-CAM \cite{eigencam}, and Layer-CAM \cite{scorecam} allows us to verify the performance of our fine-tuning task and understand the potential of our methodology. 
We use code in the repository \cite{pytorchgradcam} for this analysis. 
This study considers only fine-tuned VGG19 models, and our target layer of VGG19 in this analysis is {\tt features.36}. 
The above three algorithms of class activation mapping probe fine-tuned VGG19 models to reveal what features of cuneiform letters and their backgrounds those models capture for to identify a class. 

To visualize these results, we prepare a batch of $10$ square cropped image samples of the prescribed class, e.g., Class $1$, at first. 
The resulting images are obtained by overlaying colorized weight distribution on the above sample batch in pixel level on which a target fine-tuned VGG19 model focus for judging the prescribed class, e.g., Class $1$, on the image samples. 
All results obtained in this analysis are calculated with a fixed random seed $42$.  

\subsection{Demonstrating Fine-Tuned Models for Other Cuneiform Sentences}\label{extvalid}
\begin{figure}[htbp]
	\centering
	\begin{minipage}[b]{0.35\linewidth}
		\centering
		\includegraphics[height=3cm]{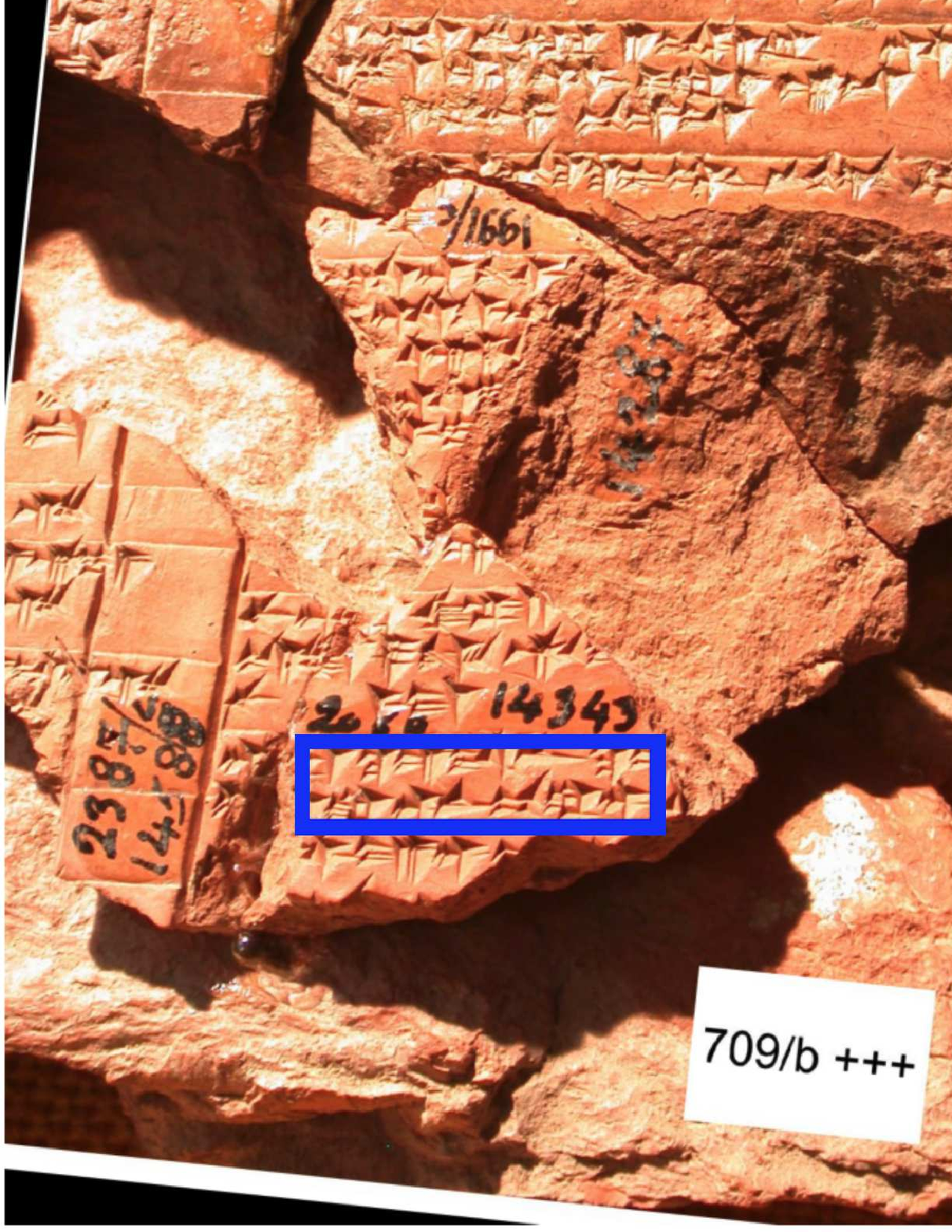}
		\label{fig:extvalidation}
	\end{minipage}
	\begin{minipage}[b]{0.35\linewidth}
		\centering
		\includegraphics[height=1cm]{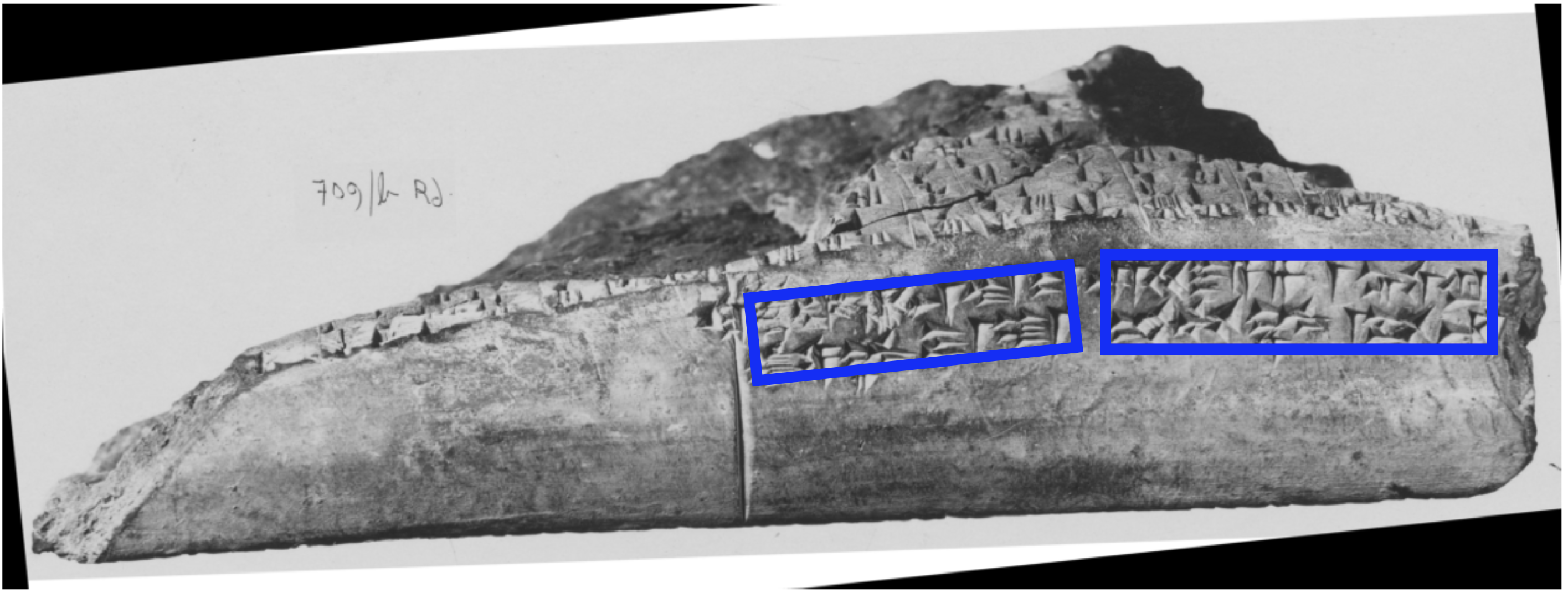}
		\label{fig:sidepic}
	\end{minipage}
\caption{Cuneiform Sentences for Studying Their Authors: (a) Right-Bottom Part of Front Side and (b) The Sides of Tablet}
\label{fig:extvalidationdata}
\end{figure}

This study is focused on writer identification analysis for other two parts of cuneiform sentences on our target tablet by applying fine-tuned models with a fixed seed of $42$. 
We borrow and use additional two pictures shown in Fig. \ref{fig:extvalidationdata} from \cite{CTH}. 
These two parts of cuneiform sentences are taken from the tablet, apart from the parts for building the main datasets: one is from the right-bottom part of the tablet, and the second one is from the side part of the tablet. 
We extract rectangular image pieces of these cuneiform sentences, as in Fig. \ref{fig:extvalidationdata} (a) and (b), and construct two datasets by cropping square images directly from their rectangular ones to apply our fine-tuned models from subsection \ref{ft}. 
After applying our models on $n$-class classification to each square image sample, we obtain scores $s_i$ using softmax function $s_i = f(x_i)$ with raw scores $r_i$ obtained from models for every class $i$ ($1\leq i \leq n$),
$$f(r_i) = \frac{e^{r_i}}{\displaystyle{\sum_{i=1}^{n} e^{r_i}}},$$
and we choose a class having the top score for the image sample. Here, $e$ is the base of natural logarithms. 
Finally, we can identify a class by 2-step majority votes.

\section{Results and Discussion}\label{Results}

\subsection{Learning Curves}\label{LCresult}
\begin{figure}[htbp]
\centering
  \begin{minipage}[b]{0.45\linewidth}
    \centering
    \includegraphics[height=3.3cm]{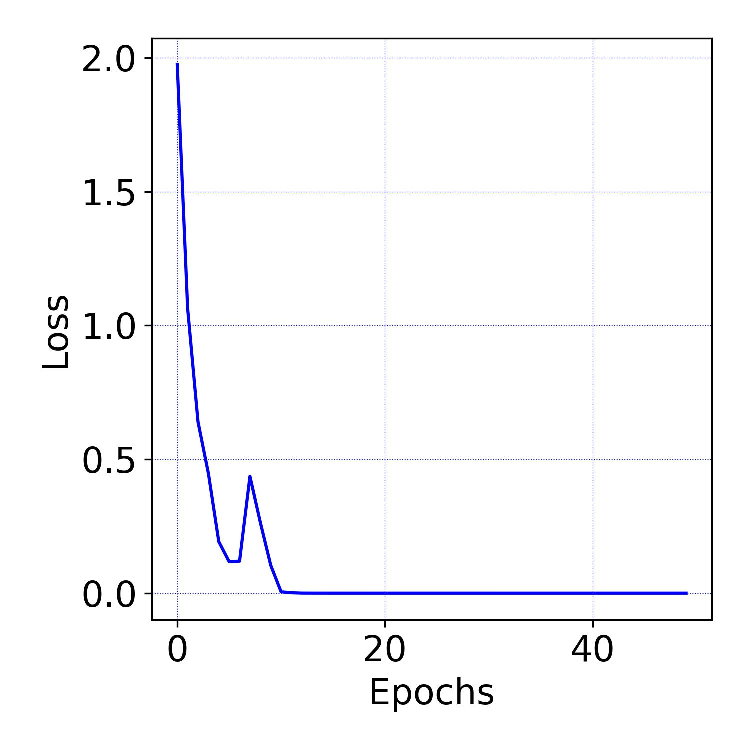}
    \label{fig:vgg19class4}
  \end{minipage}
  \begin{minipage}[b]{0.45\linewidth}
    \centering
    \includegraphics[height=3.3cm]{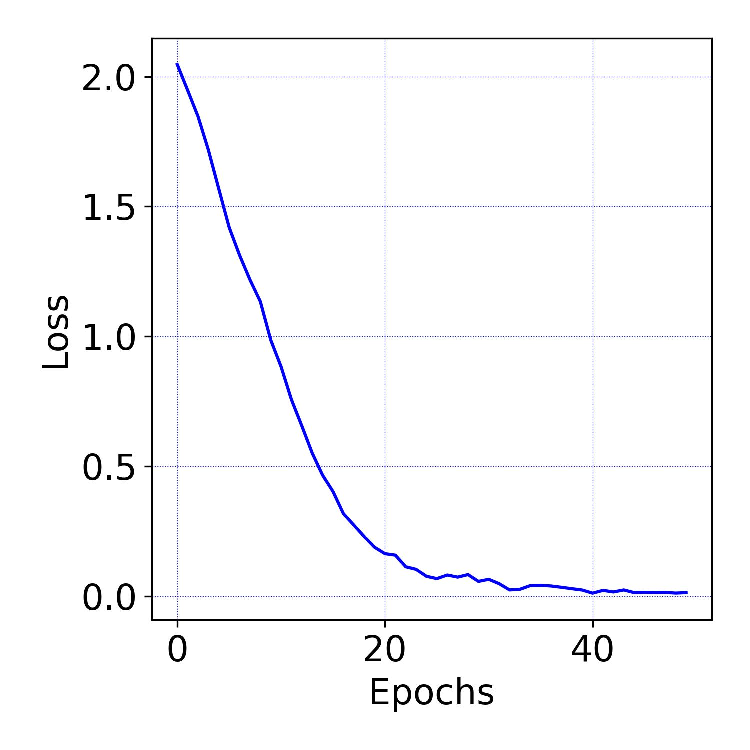}
    \label{fig:}
  \end{minipage}\medskip
  
  \centering
  \begin{minipage}[b]{0.5\linewidth}
    \centering
    \includegraphics[height=3.3cm]{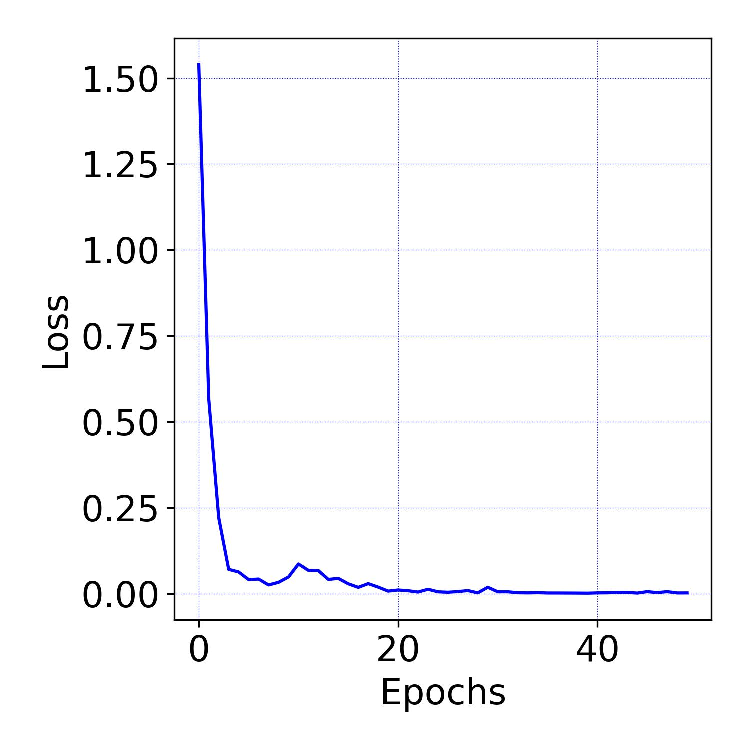}
    \label{fig:backside}
  \end{minipage}
  \caption{Learning Curves for Fine-Tuned Image Models on v001 Dataset, the Case of Seed $1033$: (a) VGG19, (b) ResNet50, and (c) InceptionV3}
  \label{fig:learningcurvesresult_4class}
\end{figure}

A learning curve is a record of loss calculated per an epoch, which demonstrates the quality of fine-tuning. 
All learning curves obtained over three models and $40$ datasets, except for the case of VGG19-based fine-tuning for v04 dataset with seed $4179$, show healthy stable convergence going to $0$ toward the end of $50$-epoch training. 
Three examples of learning curves for different pretrained models converging within $50$ epochs are shown in Fig. \ref{fig:learningcurvesresult_4class}. In each picture, the vertical axis represents the value of calculated loss at each epoch, and the horizontal axis represents the number of epochs. 

In this study, we will omit results related to the case of VGG19-based fine-tuning for v04 dataset with seed $4179$.

\subsection{Confusion Matrices}\label{CMresult}
\begin{figure}[htbp]
\centering
  \begin{minipage}[b]{0.45\linewidth}
    \centering
    \includegraphics[height=3.3cm]{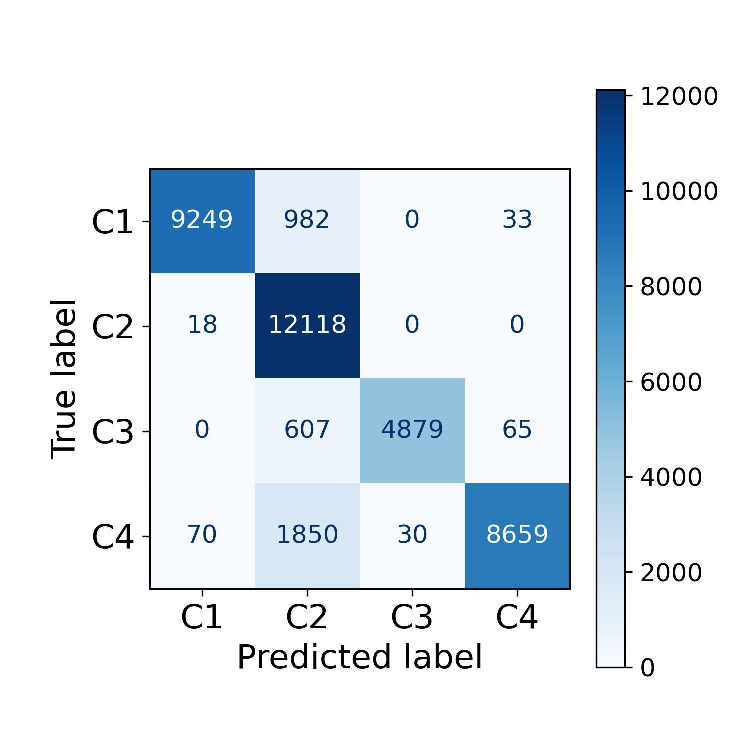}
    \label{fig:vgg19class4}
  \end{minipage}
  \begin{minipage}[b]{0.45\linewidth}
    \centering
    \includegraphics[height=3.3cm]{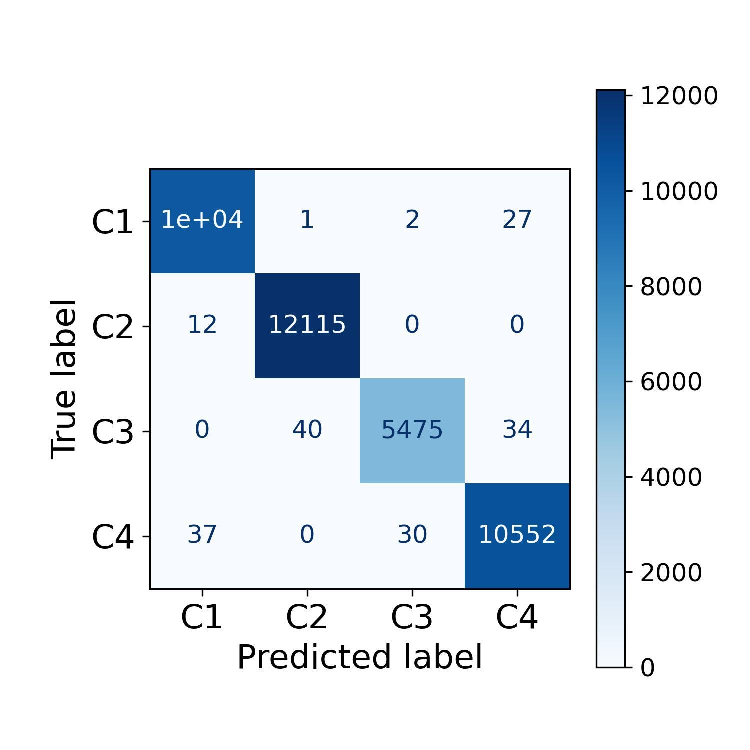}
    \label{fig:frontside2}
  \end{minipage}\medskip
  
  \centering
  \begin{minipage}[b]{0.5\linewidth}
    \centering
    \includegraphics[height=3.3cm]{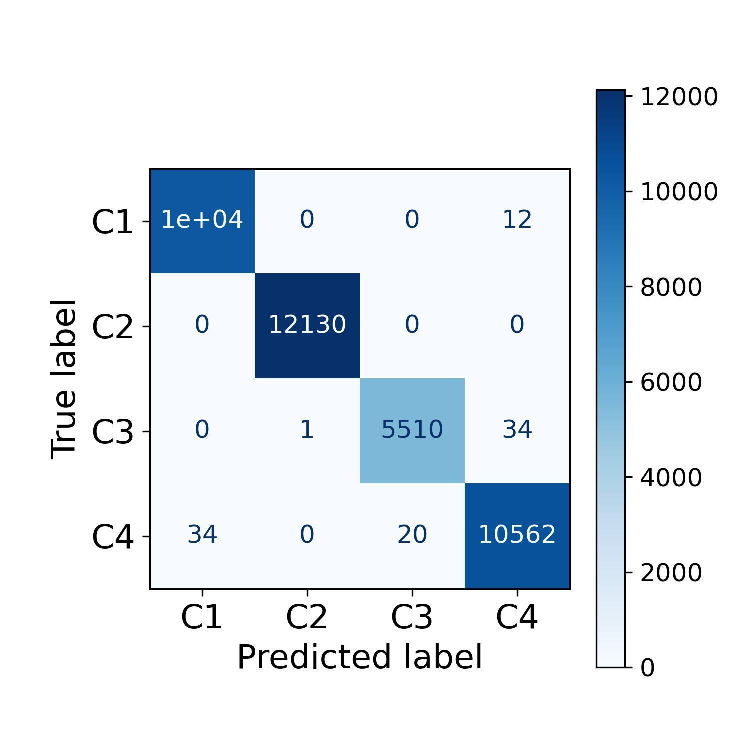}
    \label{fig:backside}
  \end{minipage}
  \caption{Overall Confusion Matrices via Fine-Tuned Models, for $4$ Classes: (a) VGG19, (b) ResNet50, and (c) InceptionV3. C$1$-C$4$ stand for Class $1$-$4$, respectively}
  \label{fig:confusionmatrixresult_4class}
\end{figure}

\begin{figure}[htbp]
\centering
  \begin{minipage}[b]{0.45\linewidth}
    \centering
    \includegraphics[height=3.3cm]{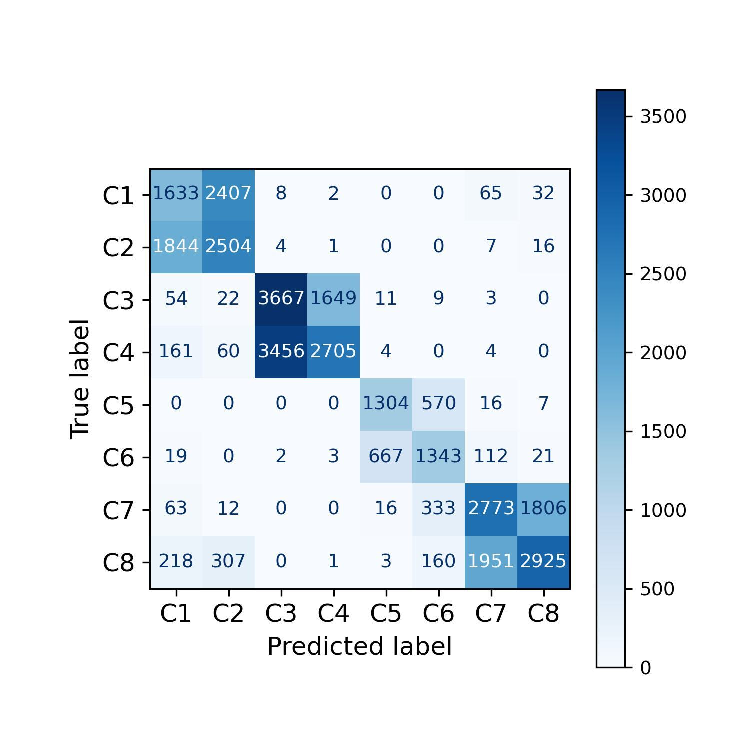}
    \label{fig:vgg19class8}
  \end{minipage}
  \begin{minipage}[b]{0.45\linewidth}
    \centering
    \includegraphics[height=3.3cm]{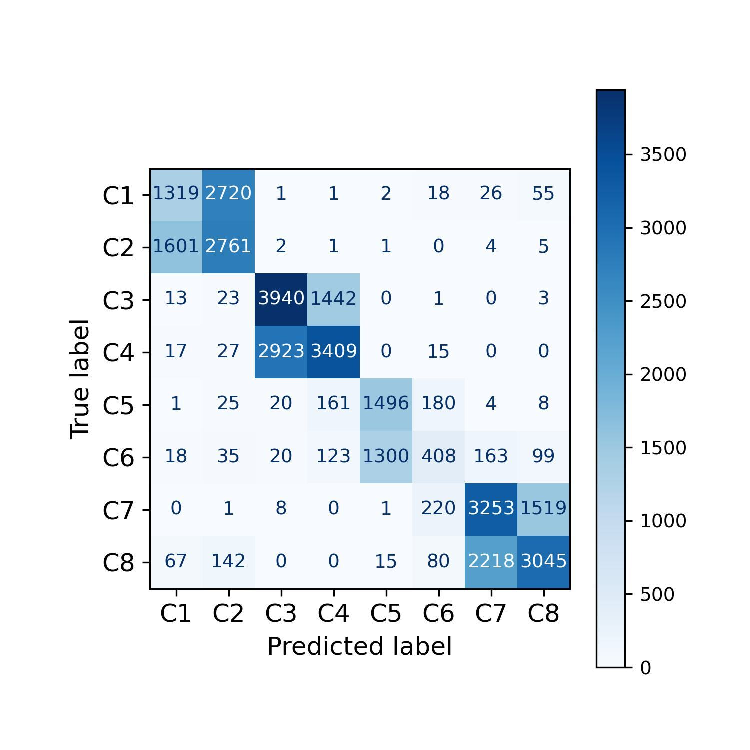}
    \label{fig:frontside2}
  \end{minipage}\medskip
  
  \centering
  \begin{minipage}[b]{0.5\linewidth}
    \centering
    \includegraphics[height=3.3cm]{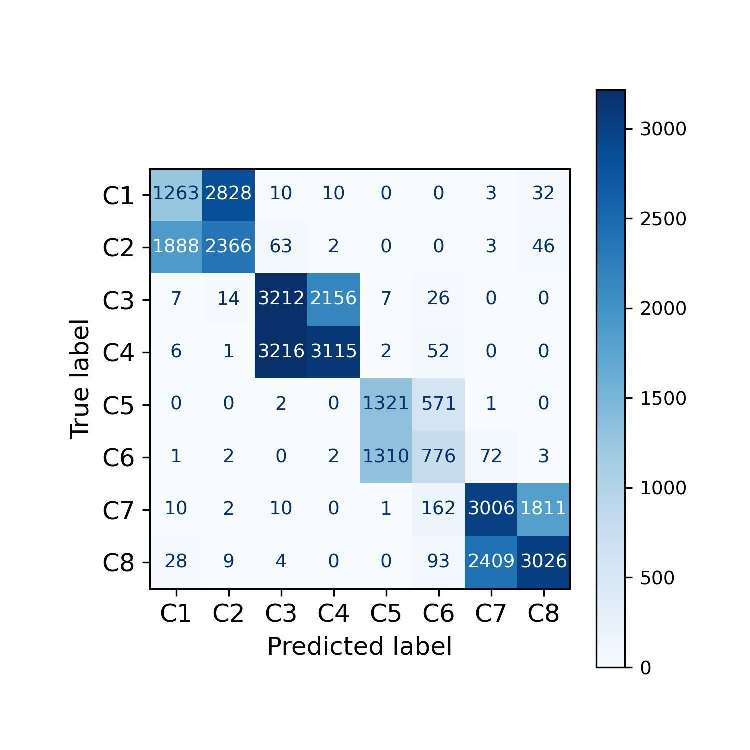}
    \label{fig:backside}
  \end{minipage}
  \caption{Overall Confusion Matrices via Fine-Tuned Models, for $8$ Classes: (a) VGG19, (b) ResNet50, and (c) InceptionV3. C$1$-C$8$ stand for Class $1$-$8$, respectively}
  \label{fig:confusionmatrixresult_8class}
\end{figure}

For each type of CNN-based image model, 
$40$ confusion matrices are obtained by applying fine-tuned models to the test part of datasets: 
$20$ for the case of $4$ classes, and $20$ for the case of $8$ classes. 
We sum up these confusion matrices, element by element, for models and classes to partially visualize the features as a snapshot in Fig.s \ref{fig:confusionmatrixresult_4class} and \ref{fig:confusionmatrixresult_8class}. 

Based on Fig.s \ref{fig:confusionmatrixresult_4class} and \ref{fig:confusionmatrixresult_8class}, we can define and compare similarities between classes as discussed in \ref{subsection:howweargue}. 
To discuss all results systematically, we introduce some useful notation: the number of samples between Class $i$ and Class $j$ in the case of $k$ classes related to the model $X$ ($X=V,R,I$ with respect to VGG19, ResNet50, and InceptionV3) as $s_{i,j}^{X,k}$ ($i \le j$). 
Its ordering can be well-defined only within the results of the same model. 
As mentioned, $s_{i,j}^{X,k}$ is the similarity between Class $i$ and Class $j$. 
The above example can be described as $s_{1,4}^{V,4} <\!\!< s_{1,2}^{V,4}$, where $s_{1,2}^{V,4} = 1000\ (=18 + 982)$ and  $s_{1,4}^{V,4} = 103\ (=33 + 70)$, and we say Class $1$ is more similar to Class $2$ than to $4$. 

Following subsection \ref{subsection:howweargue}, we denote the similarity in the case of $8$ classes related to the model $X$ between Class $i$ \& $i+1$ and Class $j$ \& $j+1$ (for $i<j$ are odd integers) as $s_{i,i+1;j,j+1}^{X,8}$. 
For instance, we see 
\begin{eqnarray}
\begin{split}
s_{1,2;7,8}^{V,8} &=
\underbrace{(65 + 32 + 7 + 16)}_{\rm right-top\ parts\ of\ the\ CM}\\
&\quad + \underbrace{(63 + 12 + 218 + 307)}_{\rm left-bottom\ parts\ of\ the\ CM}\\ 
&= 720,
\end{split}
\end{eqnarray}
from Fig. \ref{fig:confusionmatrixresult_8class} (a). 

Two-step majority vote over all confusion matrices tells us  
\begin{equation}\label{eq:3}
s_{2,3}^{X,4} \sim 0,\ {\rm and}\ \ s_{3,4;5,6}^{X,8} \sim 0.
\end{equation}
These results suggest that at least two authors wrote the present tablet. 
In addition, the fact of two repetition of the same cuneiform sentences in the tablet indicates that two authors, ``1st Author'' and ``2nd Author',' wrote this tablet. 
In the case of $4$ classes, ``1st Author'' corresponds to Classes $1$ \& $2$ and ``2nd Author'' corresponds to Classes $3$ \& $4$. 
In the case of $8$ classes, ``1st Author'' corresponds to Classes $1,2,3,$ and $4$, and ``2nd Author'' corresponds to Class $5,6,7,$ and $8$. 

A remarkable result is that we see  
\begin{eqnarray}\label{eq:1}
\begin{alignedat}{2}
s_{1,4}^{X,4} > s_{3,4}^{X,4} >\!\!> s_{1,3}^{X,4} \geq 0 \\ 
{\rm or}\ \ s_{1,4}^{X,4} \sim s_{3,4}^{X,4} >\!\!>  s_{1,3}^{X,4} \geq 0,
\end{alignedat}
\end{eqnarray}
\begin{eqnarray}\label{eq:2}
\begin{alignedat}{2}
s_{1,2;7,8}^{X,8} > s_{5,6;7,8}^{X,8} >\!\!>  s_{1,2;5,6}^{X,8} \geq 0 \\ 
{\rm or}\ \ s_{1,2;7,8}^{X,8} \sim s_{5,6;7,8}^{X,8} >\!\!> s_{1,2;5,6}^{X,8} \geq 0,
\end{alignedat}
\end{eqnarray}
for all models $X$ via 2-step majority vote. 
These results suggest that the 2nd Author's writing skills were becoming closer to the 1st Author's one. 
In other words, these results indicate the 2nd Author ``intentionally imitated'' the 1st Author. 

\subsection{Class Activation Mapping}\label{CAMresult}
\begin{figure}[htbp]
\centering
\begin{tabular}{cccccccccc}
  \begin{minipage}[b]{0.25\linewidth}
    \centering
    \includegraphics[keepaspectratio, height=4cm]{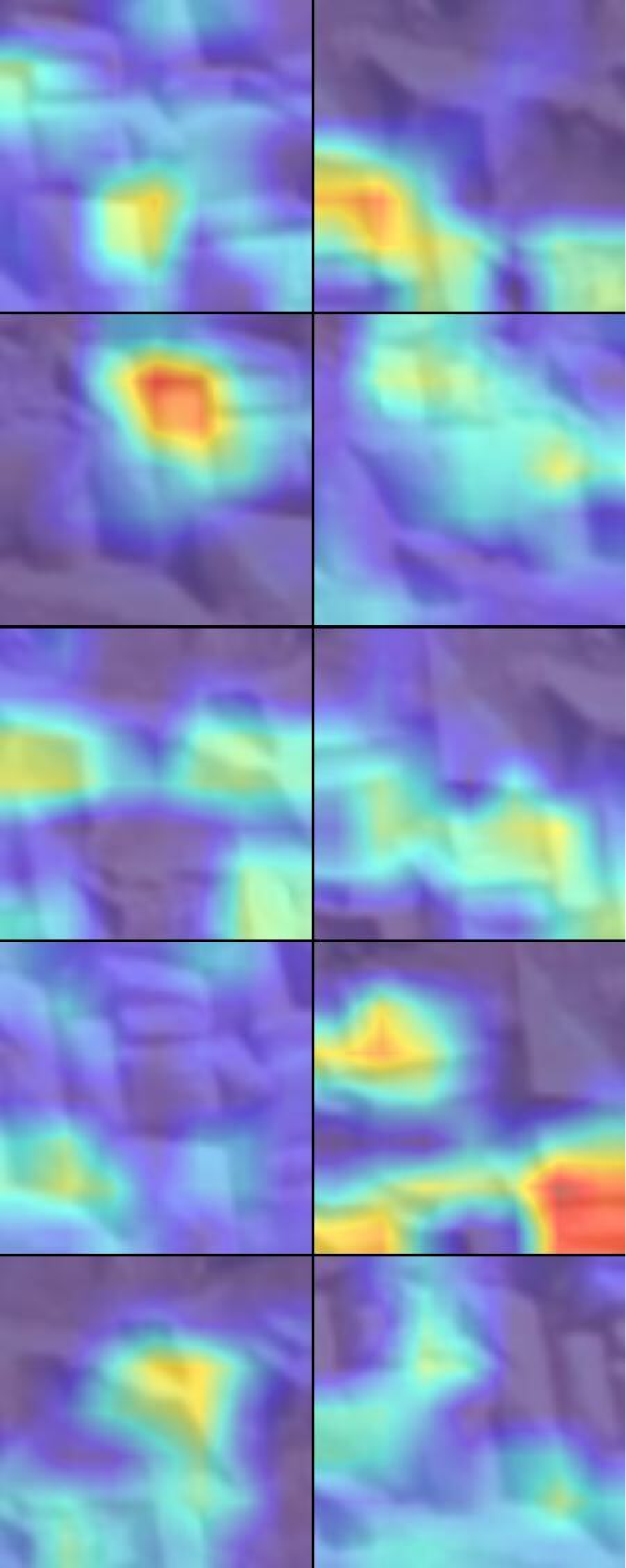}
  \end{minipage}
  \begin{minipage}[b]{0.25\linewidth}
    \centering
    \includegraphics[keepaspectratio, height=4cm]{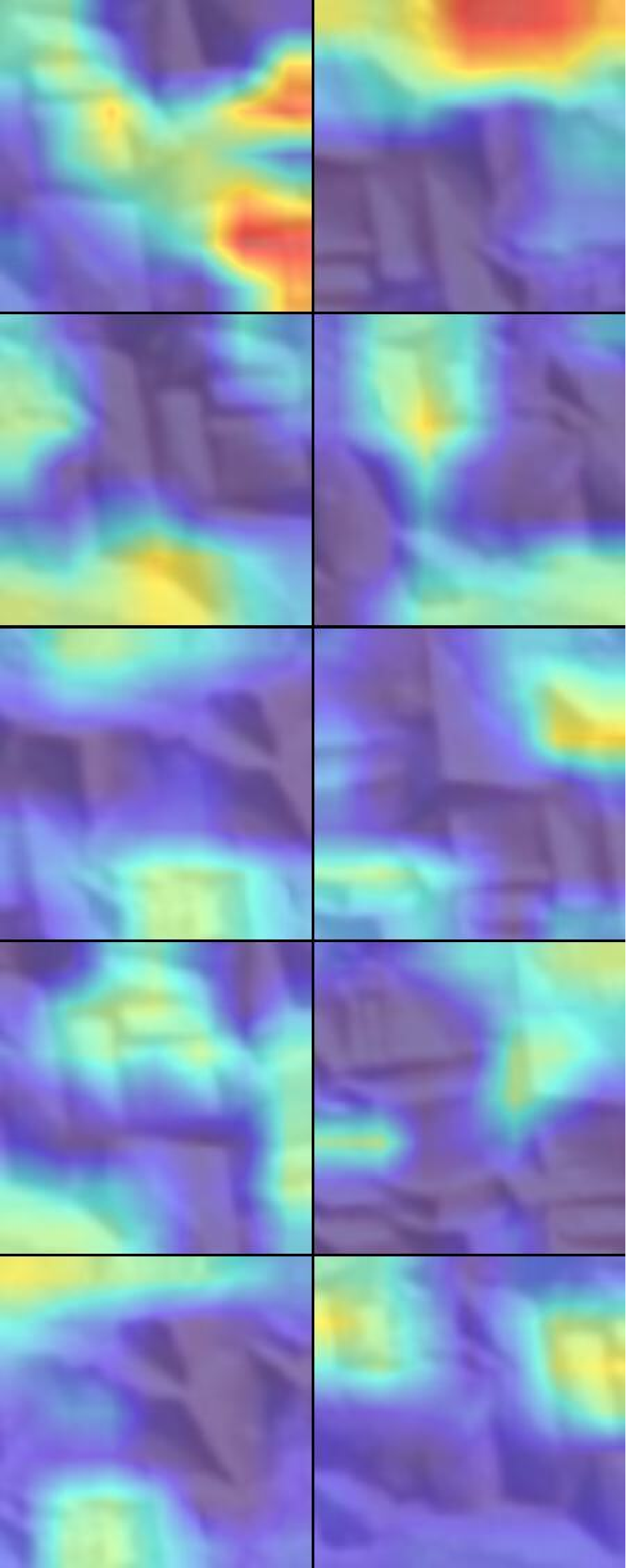}
  \end{minipage}
  \begin{minipage}[b]{0.25\linewidth}
    \centering
    \includegraphics[keepaspectratio, height=4cm]{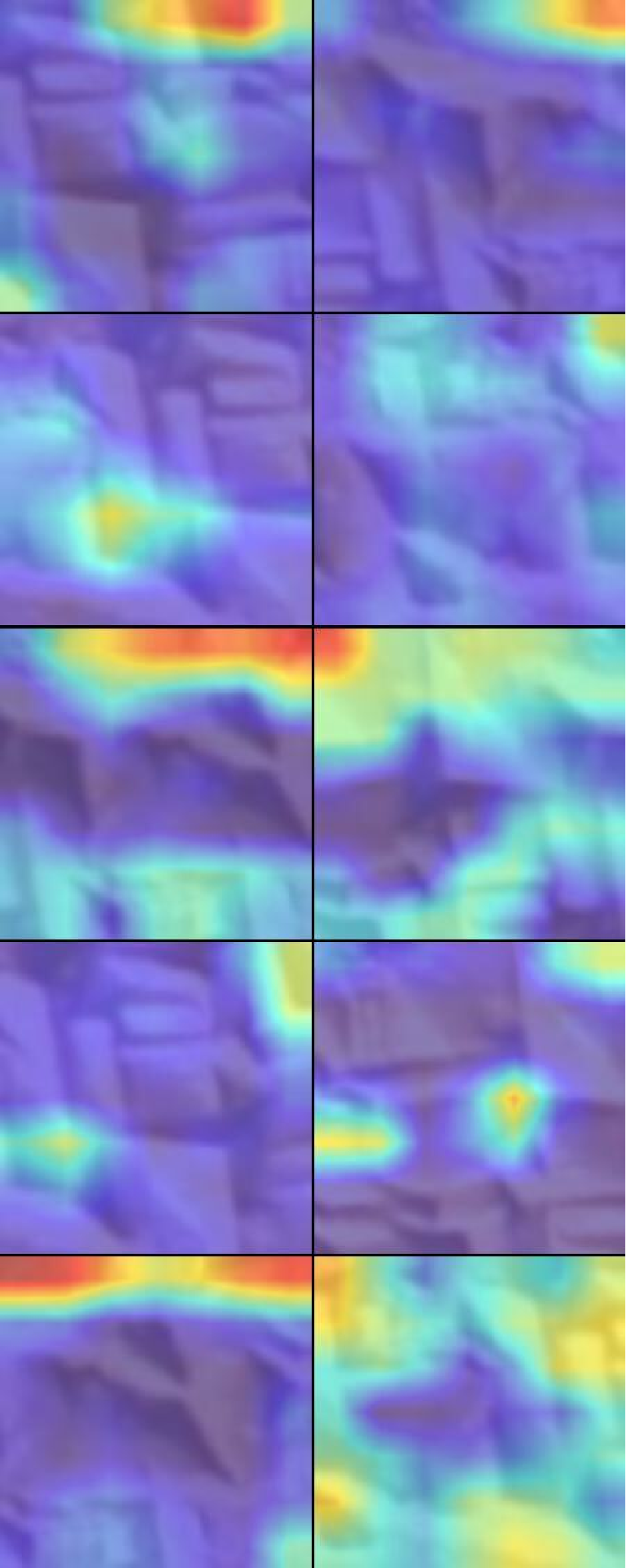}
  \end{minipage}
  \end{tabular}
\caption{Results of Class Activation Mapping for VGG19, v001 Dataset of Seed $1033$, Class $1$: (a) Grad-CAM, (b) Eigen-CAM, and (c) Layer-CAM}
\label{fig:camresult_summary}
\end{figure}

Partial results for Grad-CAM, Eigen-CAM, and Layer-CAM are shown in Fig. \ref{fig:camresult_summary}. 
Background images (their base color is purple) consist of a batch of $10$ square cropped image samples of Class $1$. 
We applied Grad-CAM, Eigen-CAM, and Layer-CAM for VGG19 models fine-tuned over v001 dataset with seed $1033$
to visualize judgement basis on predicting Class $1$ in pixel level by color temperature, as explained in subsection \ref{extvalid}. 
Red-color parts represent heavy weights used by fine-tuned models for classification, and blue-color parts represent the inverse meanings. 

Two key ingredients are found on these results: 
(1) fine-tuned VGG19 models tend to capture patterns of contours of cuneiforms rather than contours themselves, and 
(2) fine-tuned VGG19 models seem to capture blank spaces rather than contours of patterns of cuneiforms themselves. 
From (1), we see fine-tuning tasks work well. 
In addition, (1) and (2) show a possibility that the model can capture the contours of characters more clearly by devising the model construction.

\subsection{Demonstrating Fine-Tuned Models to Other Cuneiform Sentences}\label{ETresult}
\begin{figure}[htbp]
	\centering
	\begin{tabular}{cccccccccc}
		\centering
		\begin{minipage}[b]{0.43\linewidth}
			\centering
			\includegraphics[keepaspectratio, height=1.3cm]{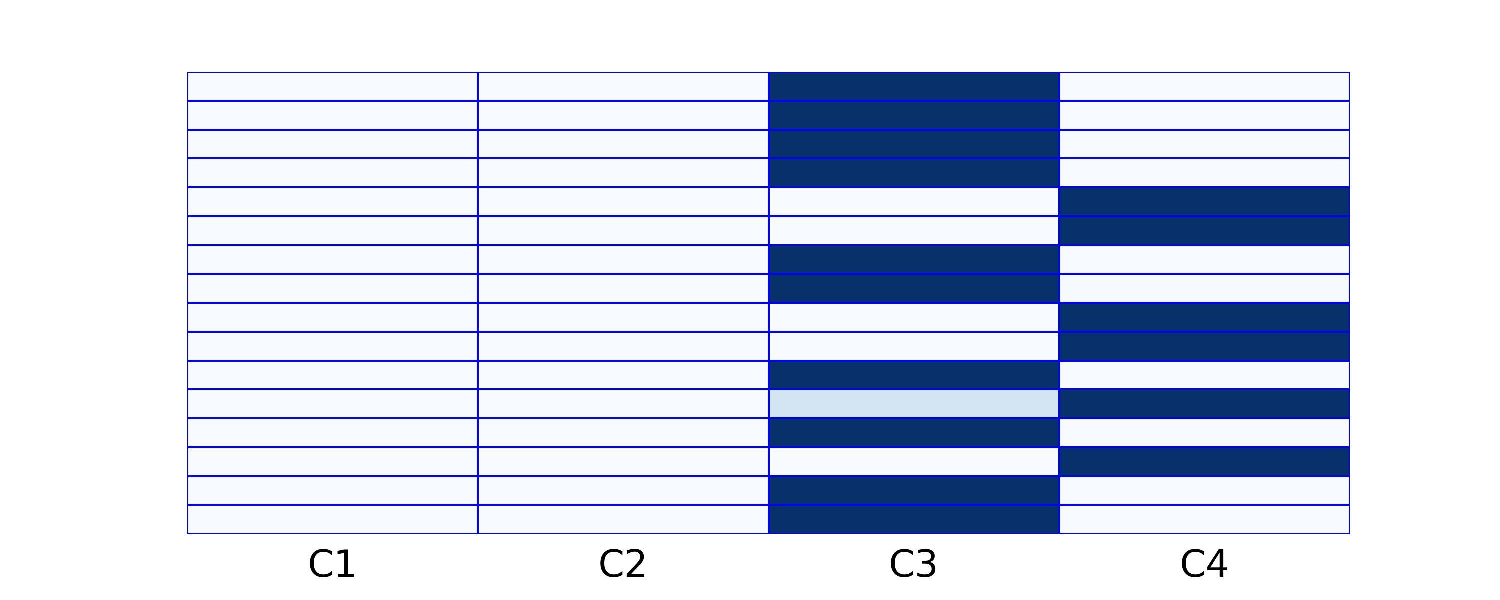}
			\label{fig:extvalidationresult_example}
		\end{minipage}
		\begin{minipage}[b]{0.43\linewidth}
			\centering
			\includegraphics[keepaspectratio, height=1.3cm]{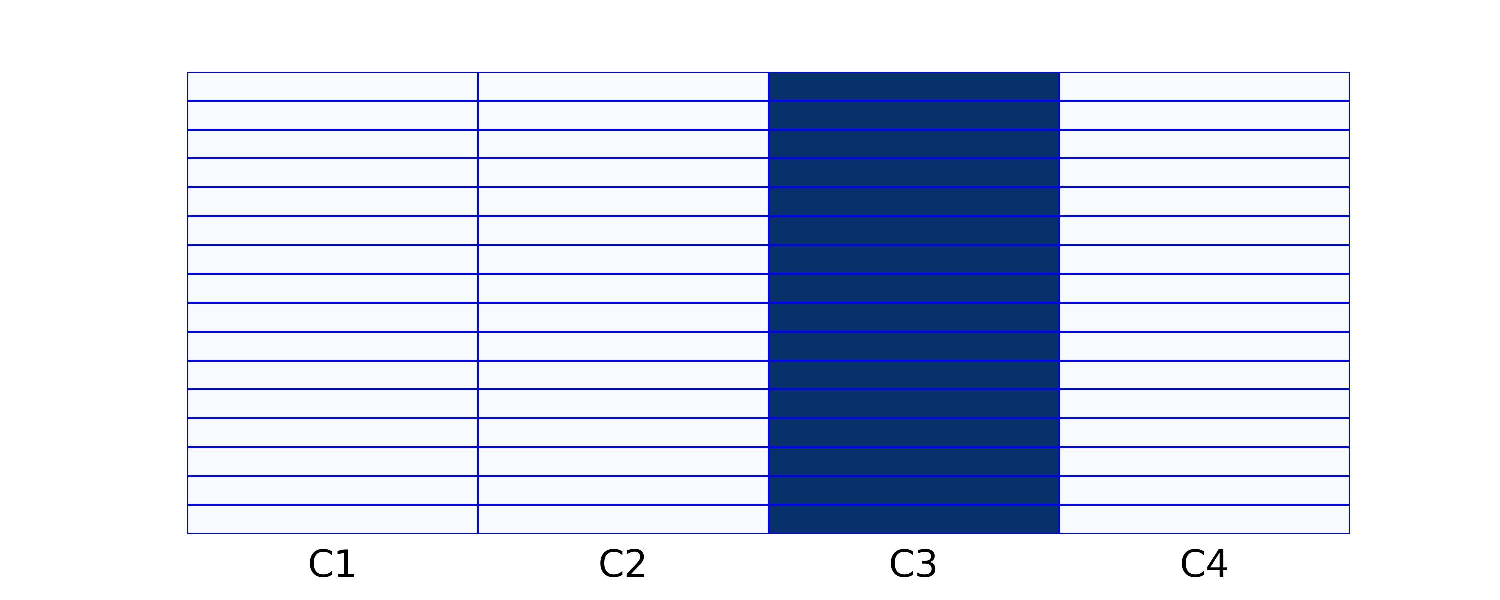}
			\label{fig:hoge}
		\end{minipage}
	\end{tabular}\medskip
	
	\centering
	\begin{tabular}{cccccccccc}
		\centering
		\begin{minipage}[b]{0.43\linewidth}
			\centering
			\includegraphics[keepaspectratio, height=1.3cm]{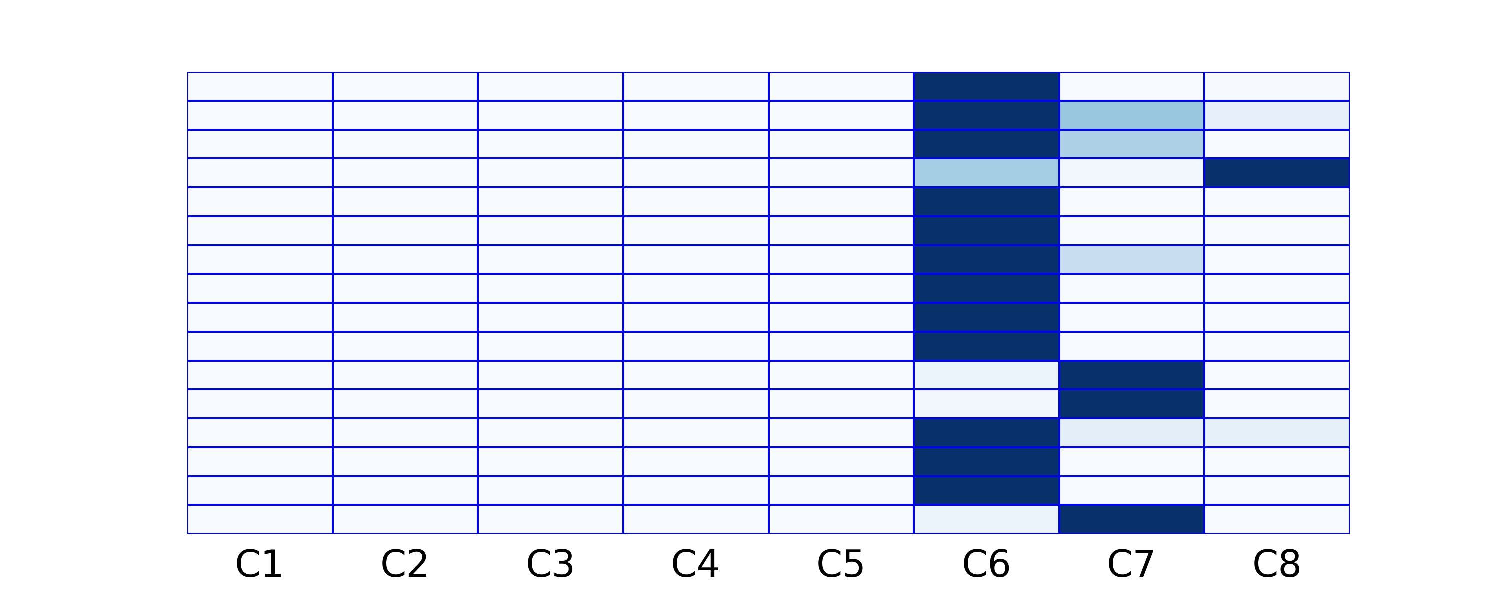}
			\label{fig:hoge}
		\end{minipage}
		\begin{minipage}[b]{0.43\linewidth}
			\centering
			\includegraphics[keepaspectratio, height=1.3cm]{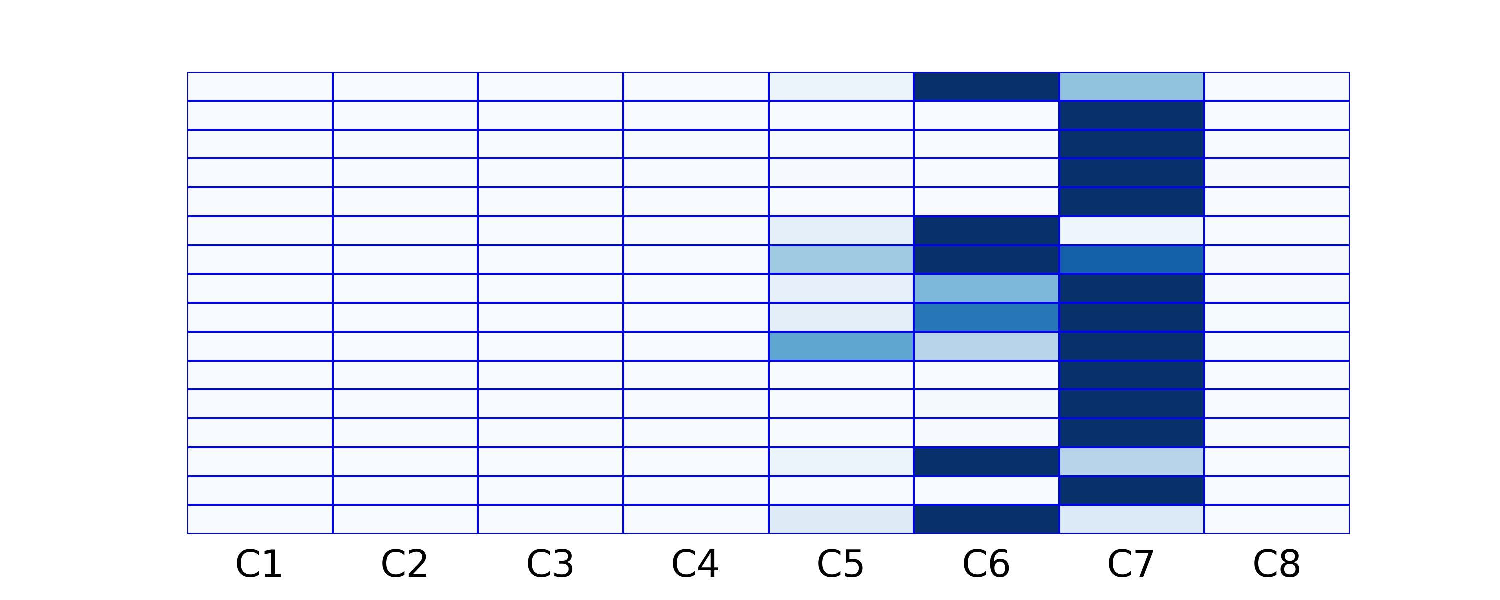}
			\label{fig:sideresult_example}
		\end{minipage}
	\end{tabular}
\caption{Scores on Testing Fig. \ref{fig:extvalidationdata} (a) By Fine-Tuned VGG19 Models: (a) D:v01,S:9325, (b) D:v01,S:2201, (c) D:v001,S:4179, and (d) D:v001,S:9325. D represents a dataset Type, and S represents a seed}
\label{fig:extvalidationresult_}
\end{figure}

\begin{figure}[htbp]
	\begin{tabular}{cccccccccc}
		\begin{minipage}[b]{0.23\linewidth}
			\centering
			\includegraphics[keepaspectratio, height=4cm]{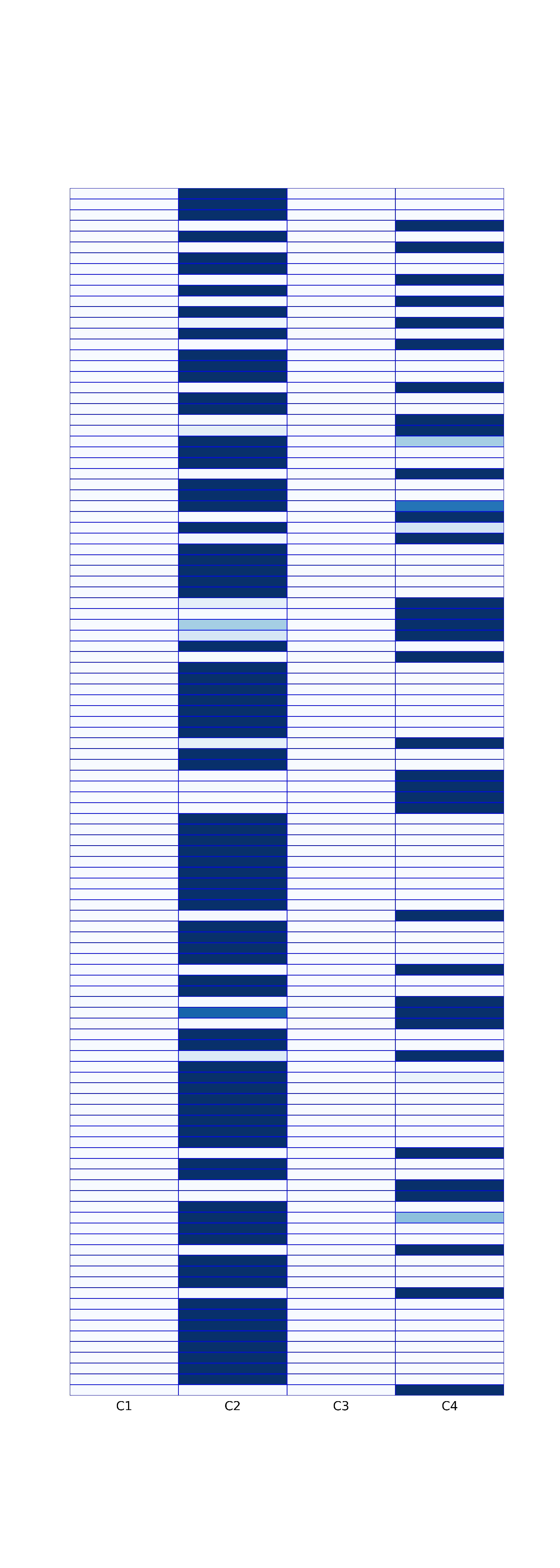}
			\label{fig:bar}
		\end{minipage}
		\begin{minipage}[b]{0.23\linewidth}
			\centering
			\includegraphics[keepaspectratio, height=4cm]{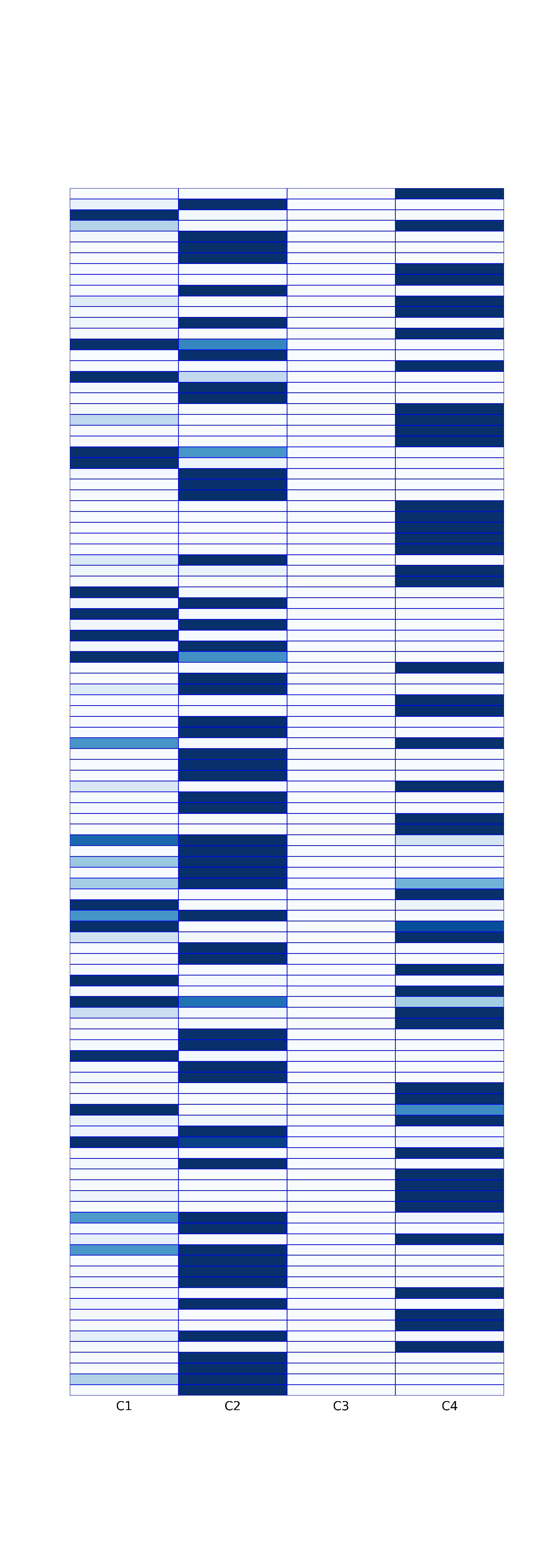}
			\label{fig:hoge}
		\end{minipage}
		\begin{minipage}[b]{0.23\linewidth}
			\centering
			\includegraphics[keepaspectratio, height=4cm]{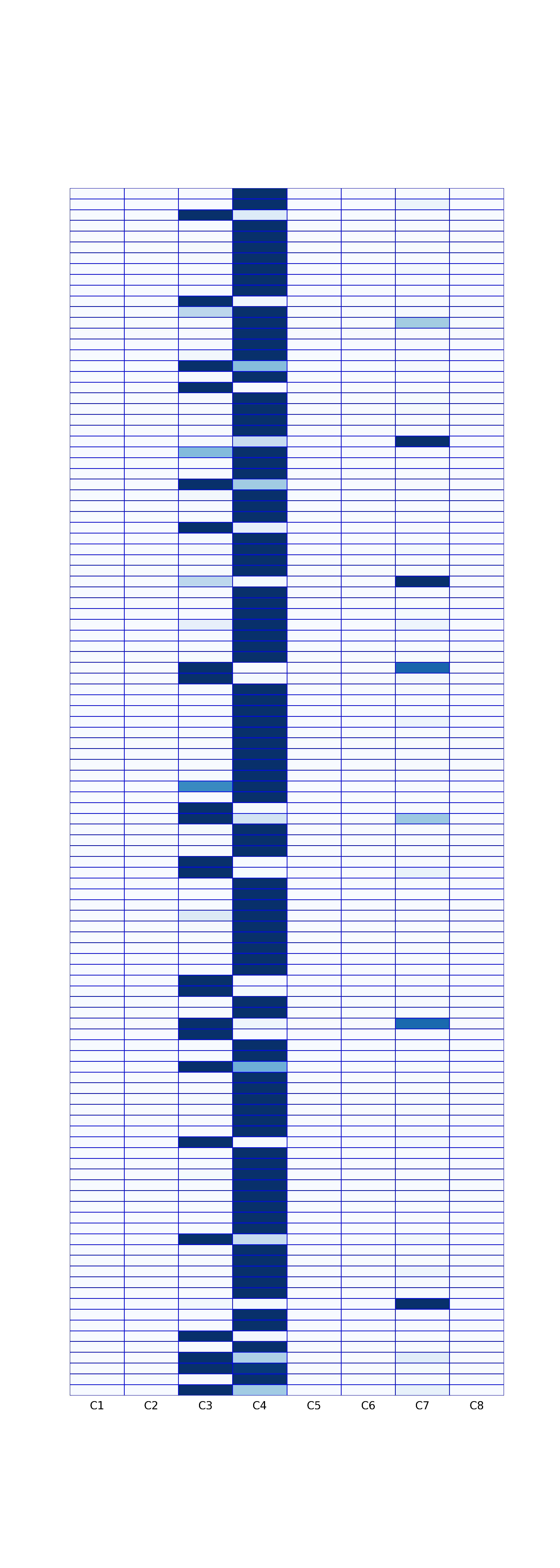}
			\label{fig:hoge}
		\end{minipage}
		\begin{minipage}[b]{0.23\linewidth}
			\centering
			\includegraphics[keepaspectratio, height=4cm]{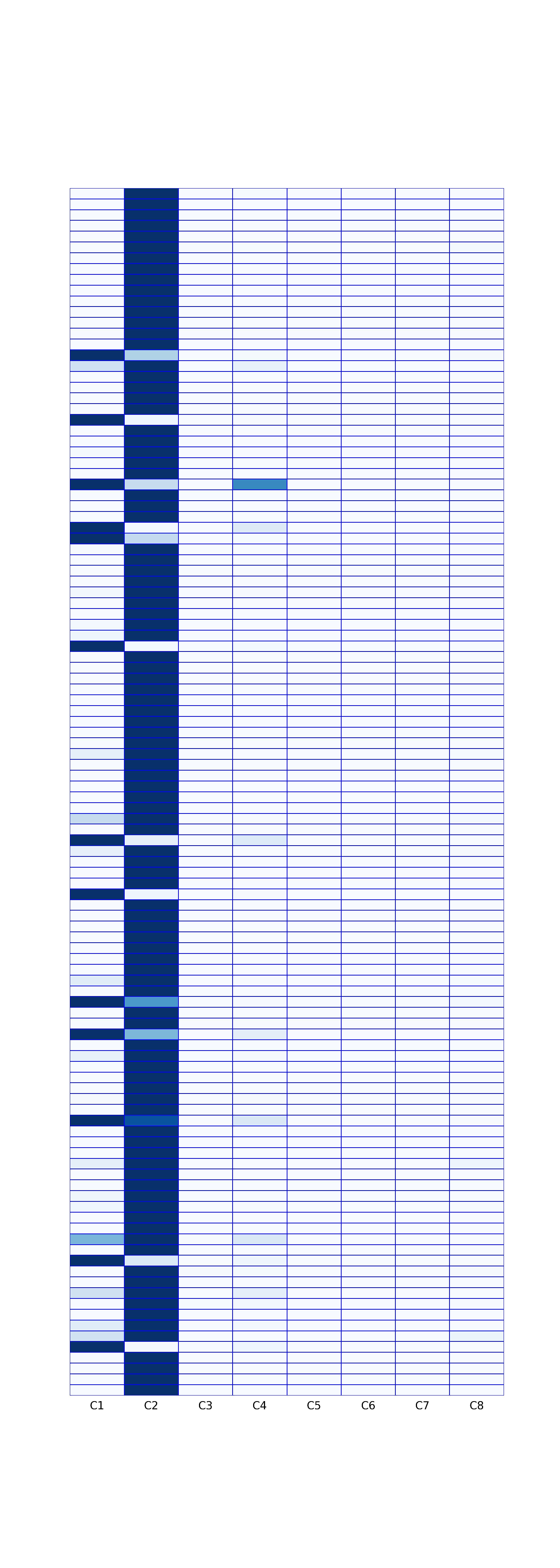}
			\label{fig:sideresult_example}
		\end{minipage}
	\end{tabular}
\caption{Scores on Testing Fig. \ref{fig:extvalidationdata} (b) By Fine-Tuned VGG19 Models: (a) D:v01,S:4179, (b) D:v01,S:9325, (c) D:v001,S:4179, and (d) D:v001,S:9325. D represents a dataset Type, and S represents a seed}
\label{fig:sideresult_}
\end{figure}

Partial results of score computations after applying our fine-tuned VGG19 models to cropped image samples of other cuneiform sentences in Fig. \ref{fig:extvalidationdata} (a) and (b) are shown in Fig.s \ref{fig:extvalidationresult_} and \ref{fig:sideresult_}, respectively. 
The vertical axis represents square image samples, the horizontal axis represents classes, and color temperature represents scores obtained by softmax function; {\it i.e.}, the darker the blue color is, the closer the score is to $1$. 
Each visualized Fig. is obtained per a triple of a fine-tuned image model, a dataset type, and a seed.

Based on the results from subsection \ref{CMresult}, 
we identify an author (either 1st Author or 2nd Author) for each of the cases of Fig. \ref{fig:extvalidationdata} (a) and (b) via 2-step majority vote, as explained in subsection \ref{extvalid}.  
A summary of results via the 1st majority vote are shown in Table \ref{table:stats_authors_extvalidation} and Table \ref{table:stats_authors_side} for the case of $4$ classes and $8$ classes, respectively. 

\begin{table}[hbtp]
  \caption{Results of 1st Majority Vote Regarding the Case of Fig. \ref{fig:extvalidationdata} (a)}
  \label{table:stats_authors_extvalidation}
  \centering
  \begin{tabular}{cccc}
    \hline
    Authors & VGG19 & ResNet50 & InceptionV3 \\
    \hline \hline
    1st Author & $2$ & $0$ & $2$ \\
    2nd Author & $38$ & $40$ & $38$ \\
    \hline
  \end{tabular}
\end{table}

\begin{table}[hbtp]
  \caption{Results of 1st Majority Vote Regarding the Case of Fig. \ref{fig:extvalidationdata} (b)}
  \label{table:stats_authors_side}
  \centering
  \begin{tabular}{cccc}
    \hline
    Authors & VGG19 & ResNet50 & InceptionV3 \\
    \hline \hline
    1st Author & $29$ & $8$ & $19$ \\
    2nd Author & $11$ & $32$ & $21$ \\
    \hline
  \end{tabular}
\end{table}

Table \ref{table:stats_authors_extvalidation} shows the 2nd Author wrote cuneiform sentences from Fig. \ref{fig:extvalidationdata} (a) via 2-step majority vote. 
This method demonstrates a potential of identifying the author of the target cuneiform sentences if a manuscript body is divided into a few pieces. 
Two-step majority vote in Table \ref{table:stats_authors_side} suggests that the 2nd Author wrote the side cuneiform sentences. 
In the case of VGG19-based fine-tuned models, the result 29:11 indicates that the author is 1st Author. On the other hand, the results of ResNet50- and InceptionV3-based fine-tuned models indicate that the author is 2nd Author. 
According to the rule of 2-step majority vote, the most likely author of the side cuneiform sentences is the 2nd Author in this study. 
This result is compatible with the current understanding of linguistics regarding this tablet, 
because these cuneiform sentences are from right-bottom parts on the front side of the tablet, where it is known the second author wrote them, indicated by the sign of his first-person writing style. 
Interestingly, 2-step majority vote leads to an appropriate answer, although separate models are alternating between the 1st and 2nd Author.

\section{Conclusion}\label{Discussions}
In this study, we analyzed clay tablet KBo 23.1 ++/KUB 30.38, which are written by two authors according to the current understanding of linguistics, by developing a new data-driven methodology based on typical CNN-based models. 
The resultant findings propose that (1) two authors, probably a ``teacher'' and a ``student,'' wrote on this tablet, (2) the second author writing the latter half of the tablet improved his writing skill as he trained, and (3) cuneiform sentences in the side part of the tablet seems to be written by the second author who wrote the latter half of the tablet. 
All results do not conflict with the current understanding of linguistics, and in particular (2) is a remarkable result that has not been previously reached by traditional linguistic methods. 
The finding (2) might indicate a certain style of teaching or education, because 
when an author wrote the back side of the tablet, he could not refer their front side. 
There might have existed a school in or around Hattusa, where this tablet were excavated, which seems to be natural because Hattusa was a capital city of the Hittite Empire. 

Crucial points of our method are the usage of newly defined notion of similarity between classes (candidate authors) and 2-step majority vote over results on augmented datasets and several CNN-based image models. 
Our proposed methodology is quite simple and extensible, and this has much potential that could be applied to linguistics/archeological problems in which related image data exists, including handwriting analysis and writer-identification analysis. 

Another important aspect of our method is that instead of focusing on optimizing one model for solving our problem, we adopt a system of majority vote over several well-trained models and several different datasets built using the same resources. 
We believe this style of modeling would become a new trend in applied research using machine learning.

\section{Supplementary information}

All detailed results in this study over main datasets and models are shown in a supplementary material. 

\section{Acknowledgments}

Authors would like to thank the authors of Catalog der Texte der Hethiter of Hethitologie-Portals Mainz \cite{CTH} on allowing the usage of samples for this study. 

\section{Code availability} 

All code for demonstrating our results is available at a GitHub repository \url{https://github.com/barrejant/Tablet_CNN_Analysis_2022}.

\bibliographystyle{sn-basic}
\bibliography{bibfile.bib}



\end{document}